%% file: main.tex
\PassOptionsToPackage{table}{xcolor}
\documentclass[10pt,twocolumn,letterpaper]{article}

 \usepackage[pagenumbers]{cvpr} % To force page numbers, e.g. for an arXiv version
\usepackage{bbm}      % \mathbbm{1}
\usepackage{dsfont}   % \mathds{1}
\usepackage{multirow}
\usepackage[pagebackref,breaklinks,colorlinks,allcolors=cvprblue]{hyperref}
\input{preamble}

\usepackage{multicol}
\usepackage{amsthm}
\usepackage[most]{tcolorbox}
\newcounter{proposition}
\crefname{proposition}{Proposition}{Propositions}
\Crefname{proposition}{Proposition}{Propositions}
\newenvironment{proposition}[1][]{%
  \refstepcounter{proposition}%
    \renewenvironment{proof}[1][\proofname]{\par\noindent{\color{proptext}\itshape ##1.}\hspace{0.5em}\ignorespaces}{\par}%
  \begin{tcolorbox}[enhanced, arc=2mm, boxrule=0.5pt,
    colframe=propframe, colback=propbody, colbacktitle=proptitle, coltitle=proptext,
    title={\textbf{Proposition~\theproposition}\ifx\relax#1\relax\else\enspace{\normalfont\color{proptext}#1}\fi},
    toptitle=2.6pt, bottomtitle=1.4pt, left=7pt, right=7pt, top=5pt, bottom=5pt, lefttitle=7pt,
    titlerule=0pt,
    segmentation code={\draw[propframe, line width=0.4pt] ([xshift=7pt]segmentation.west) -- ([xshift=-7pt]segmentation.east);}, middle=5pt,
    before skip=8pt, after skip=8pt,
    before upper={\setlength{\abovedisplayskip}{4pt}\setlength{\belowdisplayskip}{4pt}\setlength{\abovedisplayshortskip}{2pt}\setlength{\belowdisplayshortskip}{4pt}},
    before lower={\setlength{\abovedisplayskip}{2pt}\setlength{\belowdisplayskip}{3pt}\setlength{\abovedisplayshortskip}{0pt}\setlength{\belowdisplayshortskip}{3pt}}]}%
  {\end{tcolorbox}}

\newlength{\tocnumw}
\newlength{\tocsubw}
\newlength{\tocleftw}
\newlength{\tocrightw}
\newlength{\tocdigitw}
\newlength{\tocrighth}
\newsavebox{\tocrightbox}
\newcommand{\tocsec}[3][0.6em]{\par\noindent\hyperref[#2]{\makebox[\tocnumw][r]{\bfseries\color{upgreen}\ref*{#2}}\hspace{#1}#3}\par\vspace{1.5pt}}
\newcommand{\tocsub}[2]{\par\noindent\hyperref[#1]{\makebox[\tocsubw][l]{\color{upgreen}\ref*{#1}}\hspace{0.6em}#2}\par\vspace{1.5pt}}
\newcommand{\suppleft}{%
\vspace*{1pt}%
\tocsec{sec:app_reward}{Analysis of the Learnability Reward}\vfill
\tocsec{sec:app_format}{Effect of the Answer Format}\vfill
\tocsec{sec:app_o3}{Training on the Video Source of Video-Zero}\vfill
\tocsec{sec:app_molmo}{Results on Another Model Family}\vspace{-2.5pt}}
\newcommand{\suppright}{%
\vspace*{1pt}%
\tocsec{sec:app_qual}{Qualitative Results}
\tocsub{sec:app_qual_gate}{Annotations Retained and Rejected by the Teacher-Answer Gate}
\tocsub{sec:app_qual_annot}{Annotations of the MLLM Annotator during Training}
\tocsub{sec:app_qual_cmp}{Comparison with the Baselines}
\tocsec[0.55em]{sec:app_prompt}{Prompt Setting}\vspace{-2.5pt}}
\newcommand{\suppcontents}{%
\begin{tcolorbox}[enhanced, sharp corners, boxrule=0pt, frame hidden, colback=white,
  borderline north={0.8pt}{0pt}{upgreen}, borderline south={0.8pt}{0pt}{upgreen},
  boxsep=0pt, left=0pt, right=0pt, top=4.8pt, bottom=9.3pt, before skip=0pt, after skip=14pt,
  fontupper=\small, halign=left, before upper={\hypersetup{linkcolor=black}}]
\settowidth{\tocnumw}{\bfseries 10}\settowidth{\tocsubw}{10.0}\settowidth{\tocdigitw}{\bfseries 1}%
\settowidth{\tocleftw}{Training on the Video Source of Video-Zero}\addtolength{\tocleftw}{\dimexpr\tocnumw+0.6em\relax}%
\settowidth{\tocrightw}{Annotations Retained and Rejected by the Teacher-Answer Gate}\addtolength{\tocrightw}{\dimexpr\tocsubw+0.6em\relax}%
\sbox{\tocrightbox}{\begin{minipage}[c]{\tocrightw}\suppright\end{minipage}}%
\setlength{\tocrighth}{\dimexpr\ht\tocrightbox+\dp\tocrightbox\relax}%
{\normalsize\bfseries\color{upgreen}Contents}\par\vspace{4pt}
{\color{upgreen!35}\hrule height 0.4pt}\vspace{8.4pt}
{\parfillskip=0pt\noindent\hspace*{\fill}\hspace*{-\tocdigitw}\begin{minipage}[c][\tocrighth][s]{\tocleftw}\suppleft\end{minipage}\hspace*{\fill}%
{\color{upgreen!25}\vrule width 0.4pt}\hspace*{\fill}%
\usebox{\tocrightbox}\hspace*{\fill}\par}
\end{tcolorbox}}
\renewcommand\maketitlesupplementary{%
  \newpage
  \twocolumn[{%
    \centering
    \Large\textbf{\thetitle}\\
    \vspace{0.5em}Supplementary Material\\
    \vspace{1.0em}
    \normalsize\vspace{0.5\baselineskip}\suppcontents\vspace{\baselineskip}
  }]%
}

\newcommand{\note}[1]{{\rmfamily\itshape\small #1}\par\smallskip}
\newcommand{\ph}[1]{{\color{black!55}\{#1\}}}
\newtcolorbox{promptbox}[2][]{enhanced, arc=2.5mm, boxrule=0.6pt,
  colframe=upgreen, colback=white, colbacktitle=upgreen, coltitle=white,
  fonttitle=\bfseries, title={#2},
  toptitle=1.9pt, bottomtitle=0.1pt, left=5pt, right=5pt, top=3pt, bottom=3pt,
  segmentation code={\draw[upgreen!35, line width=0.3pt] ([xshift=6pt]segmentation.west) -- ([xshift=-6pt]segmentation.east);}, middle=3pt,
  fontupper=\ttfamily\footnotesize\raggedright\frenchspacing, fontlower=\ttfamily\footnotesize\raggedright\frenchspacing,
  before skip=6pt, after skip=6pt, #1}
  
\input{math_commands.tex}

\definecolor{cvprblue}{rgb}{0.21,0.49,0.74}
\definecolor{cornellred}{rgb}{0.7, 0.11, 0.11}
\definecolor{cadmiumgreen}{rgb}{0.0, 0.42, 0.24}
\definecolor{aliceblue}{rgb}{0.91, 0.94, 0.97}
\definecolor{darkblue}{rgb}{0.83, 0.89, 0.97}
\definecolor{Red7}{rgb}{0.941, 0.243, 0.243}
\definecolor{Green7}{RGB}{55, 178, 77}
\definecolor{Blue9}{rgb}{0.098,0.3,0.9}
\definecolor{darkergreen}{RGB}{21, 152, 56}
\definecolor{red2}{RGB}{252, 54, 65}
\definecolor{myteal}{RGB}{0,78,77}
\definecolor{sectiongray}{HTML}{E6E6E6}
\definecolor{mypink}{RGB}{253,231,230}

\definecolor{indomain}{HTML}{F3F9F4}
\definecolor{outdomain}{HTML}{F8F6FC}
\definecolor{upgreen}{HTML}{1B7F4B}
\definecolor{downred}{HTML}{B3261E}
\definecolor{ourspurple}{HTML}{F7F5FC}

\newcommand{\gainup}[1]{{\scriptsize\textcolor{upgreen}{+#1}}}
\newcommand{\gaindn}[1]{{\scriptsize\textcolor{downred}{$-$#1}}}

\colorlet{propframe}{upgreen!28}
\colorlet{proptitle}{upgreen!12}
\colorlet{proptext}{upgreen}
\colorlet{propbody}{indomain!60}

\colorlet{tocframe}{upgreen!28}
\colorlet{toctitle}{upgreen!12}
\colorlet{toctext}{upgreen}
\colorlet{tocbody}{indomain!60}
\usepackage{pifont}
\definecolor{cmarkgreen}{RGB}{34,139,34}
\definecolor{xmarkred}{RGB}{200,50,50}
\newcommand{\cmark}{\textcolor{cmarkgreen}{\ding{51}}}
\newcommand{\xmark}{\textcolor{xmarkred}{\ding{53}}}

\usepackage[most]{tcolorbox}

\hypersetup{
    colorlinks=true,
    linkcolor=cornellred,   
    urlcolor=Blue9,
    citecolor=cadmiumgreen
}

\def\confName{CVPR}
\def\confYear{2027}

\title{MLLM-as-an-Annotator:\\   Online Video Annotation for On-Policy Self Distillation}

\author{$\text{Jiwook Kim}^{\dagger}$\\
KAIST AI\\
{\tt\small tom919@kaist.ac.kr}
\and
$\text{Jinwoo Lee}^{\dagger}$\\
KAIST AI\\
{\tt\small bestgenius10@kaist.ac.kr}
\and
Byunghyun Jeon\\
KyungHee University\\
{\tt\small jbh030312@khu.ac.kr}
\and
Hyunjung Shim\\
KAIST AI\\
{\tt\small kateshim@kaist.ac.kr}
}

\begin{document}
\maketitle
\begingroup
\renewcommand{\thefootnote}{}
\footnotetext{$\dagger$ indicates equal contributions.}
\footnotetext{Our project page is \url{https://tom919654.github.io/MLLM_as_an_Annotator}.}
\endgroup
\input{sec/0_abstract}    
\input{sec/1_intro}

\input{sec/2_Related_Work}

\input{sec/3_Method}

\input{sec/4_Experiment}
\input{sec/5_Conclusion}
\newpage
{
    \small
    \bibliographystyle{ieeenat_fullname}
    \bibliography{main}
}

\input{sec/X_suppl}
% {
%     \small
%     \bibliographystyle{ieeenat_fullname}
%     \bibliography{supple}
% }

\end{document}

%% file: preamble.tex
\usepackage{amsmath,amssymb}
\usepackage[capitalize]{cleveref}
\input{notation}

%% file: notation.tex
\newcommand{\pis}{\pi_{\theta}}               % student
\newcommand{\pit}{\pi_{\mathrm{T}}}           % self-teacher (reads privileged evidence p^phi)
\newcommand{\pia}{\pi_{\phi}}                 % annotator (frozen MLLM + LoRA)
\DeclareMathOperator{\sg}{sg}                 % stop-gradient
 
\newcommand{\Qset}{\mathcal{Q}}               % gated questions in one step
\DeclareMathOperator{\ans}{ans}               % answer extractor (<answer> tag only)
 
\newcommand{\rbar}{\bar{r}}                   % group pass rate (1/G) sum_i r_i
 
\newcommand{\pSd}{\pis(o_{i,t}\mid\cdot,\boldsymbol{o}_{i,<t})}        % = pi_theta(o_{i,t} | v, q^phi, o_{i,<t})
\newcommand{\tlr}{\delta}                     % token log ratio teacher / student
\newcommand{\Uset}{\mathcal{U}}               % loop positions of a rollout
\newcommand{\relu}[1]{\left[#1\right]_{+}}
 
\newcommand{\gate}{g}                         % consistency gate g(q^phi) in {0,1}
\newcommand{\learn}{\ell}                     % learnability l(q^phi) = 4 rbar (1 - rbar)
\newcommand{\Rann}{R}                         % annotator reward R(q^phi)
\newcommand{\Cset}{\mathcal{C}}               % candidates of a video that pass the gate
\newcommand{\Adv}{A}                          % annotator advantage A_j
 
\newcommand{\ind}{\mathbbm{1}}                 % indicator; or \mathds{1} with \usepackage{dsfont}
 

%% file: math_commands.tex
\usepackage{amsmath,amsfonts,bm}

\def\eqref#1{equation~\ref{#1}}
\def\1{\bm{1}}

\def\va{{\bm{a}}}

\def\vo{{\bm{o}}}
\def\vp{{\bm{p}}}
\def\vq{{\bm{q}}}

\def\vv{{\bm{v}}}

\DeclareMathAlphabet{\mathsfit}{\encodingdefault}{\sfdefault}{m}{sl}
\SetMathAlphabet{\mathsfit}{bold}{\encodingdefault}{\sfdefault}{bx}{n}

%% file: sec/0_abstract.tex
\begin{abstract}
On-policy self-distillation (OPSD) post-trains a model with token-level supervision from the model itself.
However, the model serves as the teacher only when given privileged information from human-annotations.
These annotations are costly for video and are fixed before training, so the student continues to train on questions that are already answered correctly.
We propose MLLM-as-an-Annotator, an online data generation loop in which a multimodal large language model (MLLM) annotates raw videos at every training step.
The MLLM annotator generates questions, pseudo-answers, and privileged information, which replace the human annotations that OPSD requires.
The student and the MLLM annotator form a closed loop: 
the student is trained on the generated annotations, and the MLLM annotator is trained on feedback from the current student.
Without any human annotation, our method improves Qwen3-VL-4B by 3.7 points on average across 10 video benchmarks, achieves the best results on all of them.
Our method also outperforms OPSD trained on human annotations.
These results suggest that the annotations for video post-training can be learned during training rather than collected in advance.
\end{abstract}

%% file: sec/1_intro.tex
\section{Introduction}
\label{sec:1}

Large Language Models (LLMs) are now trained in multiple stages, each with a distinct goal.
Pre-training builds broad knowledge, language ability, and general representations through next-token prediction on large-scale corpora~\cite{radford2018improving, brown2020language}.
Post-training then narrows the output distribution of the pre-trained model toward behaviors useful for downstream tasks.
Post-training typically utilizes Supervised Fine-Tuning (SFT) followed by Reinforcement Learning (RL)~\cite{team2025kimi, guha2026openthoughts}.
SFT teaches the model to follow instructions and output formats from curated demonstrations.
RL further improves the model with a reward signal, such as human preference~\cite{ouyang2022training, rafailov2023direct} or verifiable correctness~\cite{shao2024deepseekmath, guo2025deepseek, team2025kimi, yu2026dapo}.
Recently, On-Policy Distillation (OPD)~\cite{agarwal2024policy, li2026rethinking} has been actively studied as an alternative to conventional RL methods such as Group Relative Policy Optimization (GRPO)~\cite{shao2024deepseekmath, guo2025deepseek}.
GRPO typically assigns one scalar reward to each sampled response. 
OPD instead provides a dense, token-level signal from a teacher model on responses sampled by the student. 
This dense signal gives finer credit assignment and lowers training cost.

\begin{figure}[t]
  \centering
  \includegraphics[width=\linewidth]{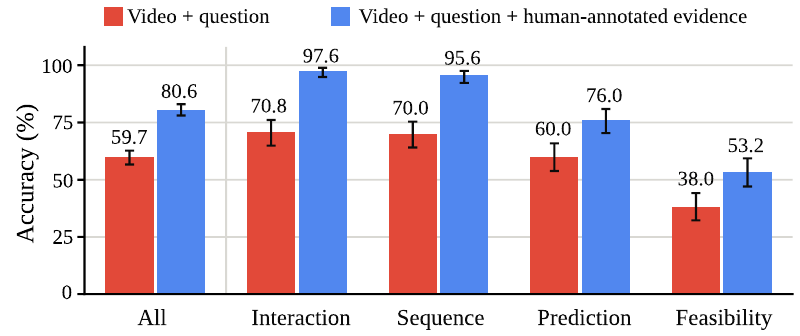}
  \vspace{-1em}
\caption{\textbf{Effect of evidence.} Accuracy of Qwen3-VL-4B on 1,000 STAR questions, with and without human-annotated evidence given in addition to the video and the question.}
  \label{fig:opsd}
  \vspace{-1.5em}
\end{figure}

Remarkably, OPD does not require the \textit{task-specific} reward design, which GRPO relies on.
Instead, OPD trains a student to minimize the reverse Kullback-Leibler (KL) divergence from the next-token distribution of a teacher on sequences sampled by the student.
This allows OPD to train on diverse tasks, which makes it a \textit{task-agnostic} post-training method.
Building on this advantage, several recent studies have improved performance across diverse tasks~\cite{xiao2026mimo, yuan2026vision, zhao2026self}.
Among these studies, On-Policy Self Distillation (OPSD) removes the external teacher model for self-improvement~\cite{zhao2026self}.
OPSD instantiates the teacher from the student model itself and gives the privileged information to the teacher, such as reference solutions.
This self-teacher provides token-level supervision on student rollouts.
As shown in \cref{fig:opsd}, the privileged information improves the accuracy of the same model, so OPSD can guide the student with the self-teacher.

However, OPSD requires additional labels, since the privileged information comes from human-annotations~\cite{zhao2026self, wang2026video}.
This limitation becomes more severe in video, where annotation requires understanding long temporal context.
The evidence for a question often lies in a few short moments, and annotators must watch the entire video to locate them.
Furthermore, Visual Question Answering (VQA) datasets are also expensive to build, since they rely on human-annotation or proprietary models such as GPT-5, Gemini, and Claude~\cite{singh2026openaigpt5card,comanici2025gemini25pushingfrontier,geminiteam2025geminifamilyhighlycapable}.

More importantly, whether a QA pair is \textit{informative} depends on the model being trained.
The model changes as training proceeds, whereas the QA pairs in existing datasets are fixed before training.
As a result, the model can continue to train on questions that are already answered correctly.
This motivates us to generate QA pairs \textit{\textbf{online}} during training.
These limitations raise a natural question:
\textit{How could we automatically generate privileged information and QA pairs that are informative for training?}

In this paper, we propose \textcolor{cornellred}{\textbf{MLLM-as-an-Annotator}}, an \textit{online data generation loop} for OPSD on video.
A Multimodal Large Language Model (MLLM) annotates raw videos at every training step, instead of building a fixed dataset.
For each sampled video, the MLLM annotator generates a question, a pseudo-answer, and privileged information that describes the visual evidence for the pseudo-answer.
The student is thus trained without the human annotations that OPSD requires.
Few video datasets provide privileged information~\cite{wu2024star,cheng2025video}, which limits OPSD on video to the domains of these datasets~\cite{wang2026video}.
In contrast, our method requires only raw videos, which cover more domains than these datasets.
As a result, the student can be trained on diverse video domains, including domains without any annotation.
Our core concept consists of three key ideas:
(1) the Teacher-Answer gate filters out inaccurate annotations,
(2) the pseudo-answer determines the \textit{direction} of distillation, and
(3) the MLLM annotator is trained \textit{online} to generate questions that are informative for the current student.
\Cref{fig:motiv} illustrates the motivation of each key idea.

First, the Teacher-Answer gate discards inaccurate annotations before training the student.
The MLLM annotator can generate a wrong pseudo-answer or privileged information that does not support the pseudo-answer, and such annotations hinder the training of the student.
The Teacher-Answer gate retains an annotation only if the self-teacher, given the privileged information, reproduces the pseudo-answer.
The student is thus trained only on annotations that the self-teacher confirms, without any human verification.

Second, the pseudo-answer determines the \textit{direction} of distillation.
In OPSD, the self-teacher alone determines this direction, regardless of whether a student rollout is correct.
OPSD can therefore encourage tokens of incorrect rollouts and penalize tokens of correct ones.
We instead verify each rollout against the pseudo-answer to set the direction, and use the self-teacher only to weight the tokens within each rollout.
As a result, the student learns toward correct rollouts and away from incorrect ones.

Third, the MLLM annotator is updated online to generate \textit{informative} questions.
We reward the MLLM annotator for questions that the current student answers correctly in some rollouts and incorrectly in others~\cite{huang2026r,foster2026learningreasonfrontierlearnability,bae2026online}.
Such questions are \textit{informative}, as they yield correct rollouts to encourage and incorrect ones to penalize.
The student and the MLLM annotator thus form a \textit{closed loop}: the student is trained on the generated annotations, and the MLLM annotator on feedback from the student.
In contrast to a fixed dataset, the training data are generated for the current student.

We evaluate MLLM-as-an-Annotator on 10 video benchmarks that cover general video understanding, temporal perception, spatial reasoning, and complex reasoning.
Our method improves the average of Qwen3-VL-4B from 48.2 to 51.9 without human annotations, and achieves the best results on all 10 benchmarks.
Our method also outperforms Video-OPSD~\cite{wang2026video}, which trains on human annotations, and Video-Zero~\cite{zhang2026video}, which trains on self-generated questions with GRPO.
The gains hold with Qwen3-VL-8B, where our method improves the average from 51.6 to 53.2.

Our core contributions are summarized as follows:
\begin{tcolorbox}[
  colback=gray!8,
  colframe=myteal,
  boxrule=0.6pt,
  arc=3mm,
  left=2.5mm,right=2.5mm,top=2mm,bottom=2mm
]
\begin{itemize}[label=\raisebox{0.2ex}{\tiny$\bullet$}, leftmargin=*, itemsep=2pt, topsep=2pt]
  \item We propose MLLM-as-an-Annotator, an \textit{online} data generation loop for OPSD on video. OPSD thus applies to raw videos without human annotations.

  \item
  We introduce the Teacher-Answer gate to retain reliable annotations, and determine the direction of each update from the pseudo-answer and its magnitude from the self-teacher. We further train the MLLM annotator online to keep positive and negative signals balanced for the current student.

  \item
  On 10 video benchmarks, our method improves the average of Qwen3-VL-4B from 48.2 to 51.9 and Qwen3-VL-8B from 51.6 to 53.2. Our method also outperforms OPSD trained on human annotations.
\end{itemize}
\end{tcolorbox}

%% file: sec/2_Related_Work.tex
%% ===================== Section 2 =====================

\begin{figure*}[th!]
  \centering
  \includegraphics[width=\linewidth]{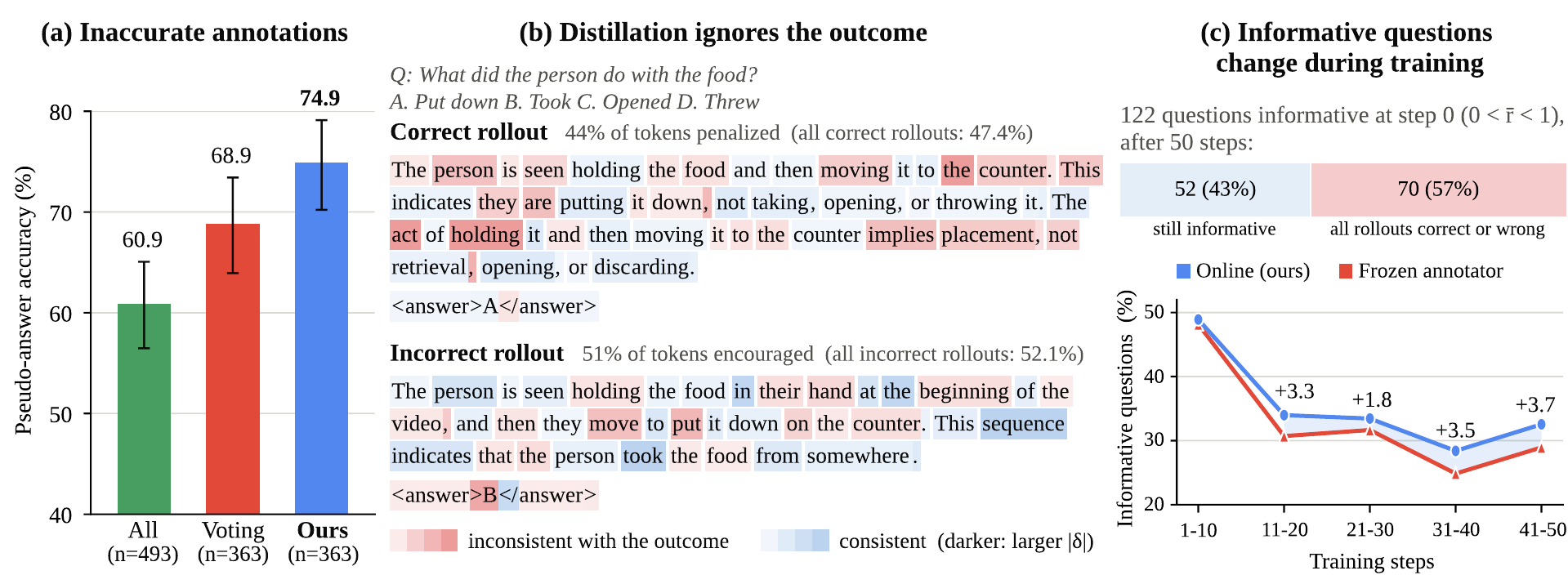}
  \vspace{-1.0em}
  \caption{\textbf{Motivations.}
(a) Accuracy of the pseudo-answers from the MLLM annotator, measured on 493 questions with human-annotated answers from STAR~\cite{wu2024star}, NExT-QA~\cite{xiao2021next}, Perception Test~\cite{patraucean2023perception}. The Teacher-Answer gate retains 363 pseudo-answers with 74.9\% accuracy, compared with 60.9\% before the gate and 68.9\% for a VisPlay-style~\cite{He_2026_CVPR} voting filter at the same pass rate.
(b) Teacher--student log ratio $\delta$ of each token for a correct and an incorrect rollout of the same question. In both rollouts, about half of the tokens have a sign of $\delta$ that is inconsistent with the correctness of the rollout.
(c) Top: The frozen annotator generates 122 questions that are \textit{informative} for the initial student, but only 52 remain informative after 50 training steps.
Bottom: Percentage of informative questions generated by a frozen and an online annotator.
The online annotator generates more informative questions.
}
  \label{fig:motiv}
  \vspace{-1em}
\end{figure*}

\section{Related Work}
\label{sec:2}

\subsection{Video Self-Distillation}
\label{sec:2.1}

\paragraph{On-policy self-distillation.}
On-policy distillation~\cite{agarwal2024policy} trains a student on self-generated samples with token-level supervision from a larger teacher model.
On-policy self-distillation (OPSD)~\cite{zhao2026self,hubotter2026reinforcement} does not require a larger model, since the student itself serves as the teacher.
This self-teacher receives privileged information that is not given to the student, and scores each token of the sampled response, which provides dense supervision.
The privileged information is a reference solution for text~\cite{zhao2026self}, and human-annotated evidence frames with the answer~\cite{wang2026video} or a clue interval~\cite{wang2026look} for video.
However, all of these methods annotate privileged information for every question before training.
OPSD is therefore limited to annotated datasets with a fixed set of questions and cannot be applied to unlabeled videos.
We instead generate the questions and the privileged information with an MLLM annotator trained online with the student, so that OPSD applies to unlabeled videos.

\paragraph{On-policy distillation with verifiable rewards.}
Distillation alone can penalize tokens of correct rollouts and encourage tokens of incorrect rollouts, because the sign of the teacher--student log ratio does not depend on whether the rollout is correct~\cite{lin2026policy}.
Recent work therefore combines distillation with verifiable rewards, so that the reward determines the direction of each update and the teacher only the token-level magnitude~\cite{lin2026policy,yang2026self,liu2026self,li2026unifying}.
OPDVR~\cite{lin2026policy} uses the direction from the reward with a teacher trained by reinforcement learning, and RLSD~\cite{yang2026self} with a privileged self-teacher.
VISD~\cite{lin2026visd} applies the direction from the reward to video reasoning with an external judge model.
However, all of these methods compute the reward from human-annotated answers.
The direction from the reward is therefore unavailable for unlabeled videos, which have no human-annotated answers.
We instead determine the direction from the pseudo-answer generated by the MLLM annotator, which requires no human-annotated answer.

\subsection{Self-Evolving Multimodal Models}
\label{sec:2.2}

\paragraph{Questioner--solver loops.}
Self-evolving methods train a model on questions that the model generates from unlabeled data, first for text~\cite{huang2026r,zhao2026absolute,sick2026onpolicyselftrainingdatafreeselfimprovement}, then for images~\cite{He_2026_CVPR,wang2026v} and videos~\cite{zhang2026video,huang2026evovid,zhu2026tempo}.
A questioner generates the questions, and a solver learns to answer them.
Both roles are initialized from the same base model and updated in alternation.
For videos, Video-Zero~\cite{zhang2026video} organizes the loop around temporally localized evidence, and EvoVid~\cite{huang2026evovid} adds rewards based on temporal order and segment localization.
In both methods, however, the solver must predict the evidence as a temporal span, and this prediction yields a single reward per rollout.
This reward does not indicate which tokens lead to a correct or incorrect answer.
We instead give the evidence to the self-teacher as privileged information, so the self-teacher provides token-level supervision.

\paragraph{Labels and learning signal.}
In questioner--solver loops, the questioner is rewarded for questions on which the sampled answers of the solver disagree~\cite{huang2026r,He_2026_CVPR,zhang2026video}, and the solver is trained with GRPO~\cite{shao2024deepseekmath} on majority-vote labels~\cite{He_2026_CVPR,wang2026v,zhang2026video}.
However, a majority-vote label is the answer that the solver already prefers.
When this answer is incorrect, GRPO increases the probability of the incorrect answer.
Tempo-Zero~\cite{zhu2026tempo} instead renders videos by executing programs, which also produce the answer to every question.
These labels do not depend on the solver, but the videos are limited to rendered ones.
Our pseudo-answer is generated by the MLLM annotator for raw videos, without majority votes or executable programs.
The Teacher-Answer gate retains the annotation only if the self-teacher, given the privileged information, reproduces the pseudo-answer.

%% file: sec/3_Method.tex
%% ===================== Section 3 =====================

\begin{figure*}[th!]
  \centering
  \includegraphics[width=\linewidth]{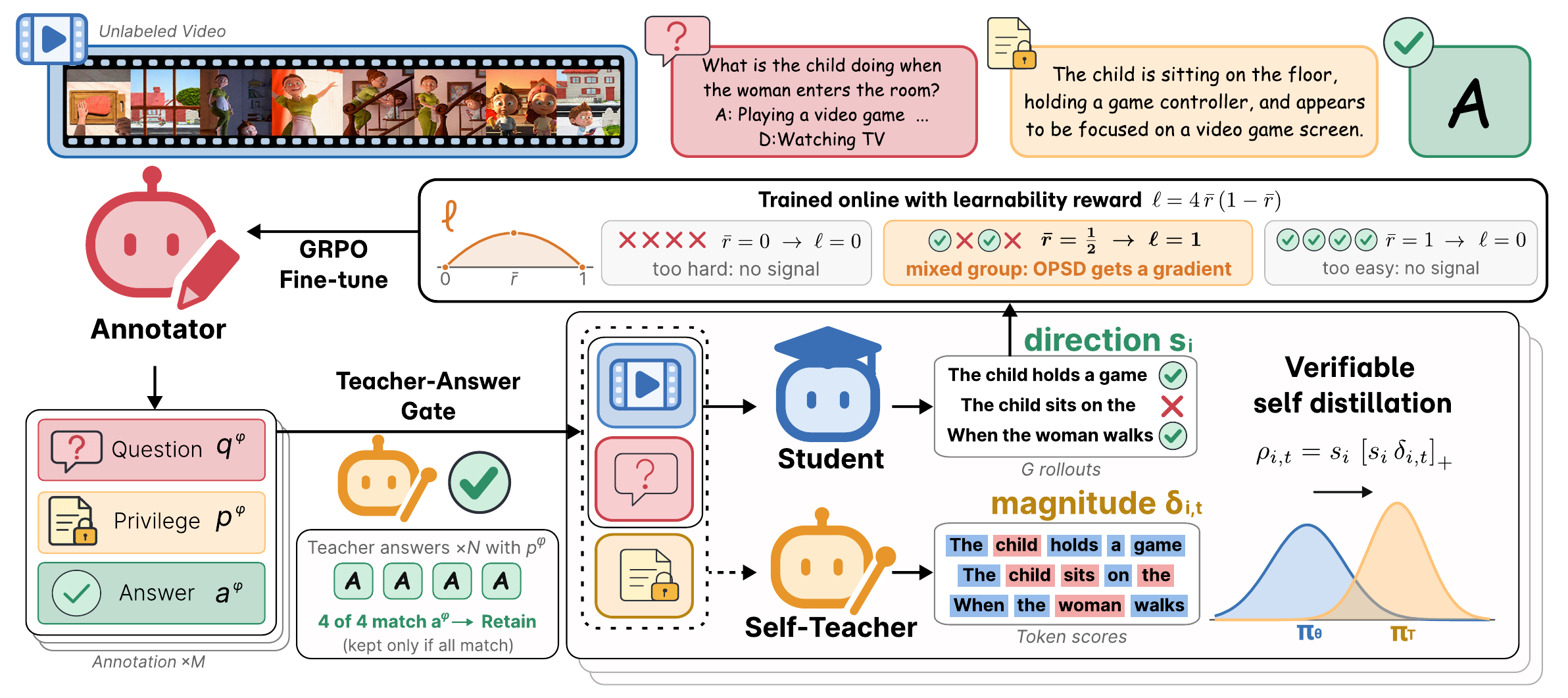}
  \vspace{-1.0em}
  \caption{\textbf{Overview.}
  For each unlabeled video $\vv$, the MLLM annotator $\pia$ generates a question $\vq^{\phi}$, a pseudo-answer $\va^{\phi}$, and privileged information $\vp^{\phi}$.
  The gate retains an annotation only if the self-teacher reproduces $\va^{\phi}$.
  The student samples $G$ rollouts from the video and the question alone.
  The pseudo-answer $\va^{\phi}$ sets the direction $s_i$ of each rollout, and the self-teacher sets the magnitude $\tlr_{i,t}$ of each token.
  Under the joint condition, the student is distilled only on tokens whose magnitude has the same sign as the direction.
  The same rollouts give the learnability reward $\learn$ that trains the MLLM annotator online.}
  \label{fig:overview}
  \vspace{-1em}
\end{figure*}

\section{MLLM-as-an-Annotator}
\label{sec:3}

MLLM-as-an-Annotator aims to train a student on diverse VQA tasks without human-annotation or a stronger model.
The student model is trained solely with one base model, because the MLLM annotator, student, and self-teacher share a same model at the beginning of training.
We utilize OPSD for task-agnostic post-training, as OPSD does not require any task-specific reward design.
OPSD instead requires privileged information $\vp^{\phi}$, which the MLLM annotator generates with questions $\vq^{\phi}$ and pseudo-answers $\va^{\phi}$.

However, training OPSD with these annotations raises three challenges as \cref{fig:motiv}.
First, the annotations can be inaccurate, which hinders the training of the student.
We introduce Teacher-Answer gate, which accepts an annotation only when the self-teacher reproduces the pseudo-answer $\va^{\phi}$ as \cref{sec:3.1}.
Second, OPSD can penalize tokens of correct rollouts and encourage tokens of incorrect rollouts.
We let the pseudo-answer $\va^{\phi}$ determine whether each rollout is reinforced or penalized, and use the self-teacher only to weight the tokens (\cref{sec:3.2}).
Third, the questions that are informative for the student change during training, and the MLLM annotator must follow this change.
We train the MLLM annotator online with a reward computed from these verification results (\cref{sec:3.3}).
The reward favors questions on which the current student produces both correct and incorrect rollouts.
\Cref{fig:overview} shows our overall pipeline, in which the student and the MLLM annotator form a closed loop.

\subsection{Annotation Verification}
\label{sec:3.1}
In this section, we introduce how to filter out unreliable annotations before training the student.
Annotations with incorrect pseudo-answers lower the reliability of the generated data, which hinders the training.
In OPSD, the self-teacher receives the description of the visual evidence as privileged information, and the student learns the next-token distributions of the self-teacher as \cref{sec:2.1}.
The privileged information for an incorrect pseudo-answer can contain incorrect evidence, which guides the self-teacher toward the incorrect answer.
The student then learns this incorrect answer from the self-teacher.
However, raw videos provide no human-annotation to detect incorrect pseudo-answers.
We therefore introduce gating, which retains an annotation only if the self-teacher reproduces the pseudo-answer from its description.
This gating improves the reliability of the generated data without any human-annotation (\cref{fig:motiv}-(a)). 
The self-teacher receives only descriptions from which it reproduces the pseudo-answer, so its next-token distributions guide the student toward this answer.

\paragraph{Annotation generation.}
The MLLM annotator $\pia$ samples $M$ candidate annotations for each video $\vv$.
Each candidate consists of a question $\vq^{\phi}$, a pseudo-answer $\va^{\phi}$, and privileged information $\vp^{\phi}$, generated together:
\begin{equation}
(\vq^{\phi},\,\va^{\phi},\,\vp^{\phi})\sim\pia(\cdot\mid\vv).
\label{eq:annot}
\end{equation}
We restrict the questions to multiple-choice or numerical formats over predefined question types.
These formats are standard in RL-based post-training~\cite{guo2025deepseek,feng2025videor1reinforcingvideoreasoning}, since their answers can be verified by exact matching.
We use the same exact matching to compare the answer of each student rollout with $\va^{\phi}$.
The privileged information $\vp^{\phi}$ describes the visual evidence for the pseudo-answer and explains why this evidence supports it.
The self-teacher receives $\vp^{\phi}$ in addition to the video and the question, while the student receives only the video and the question.

\paragraph{Teacher-Answer gate.}
The Teacher-Answer gate filters out unreliable annotations before distillation.
We regard an annotation as reliable if its privileged information leads the self-teacher to the pseudo-answer.
Therefore, we sample $N$ responses $\tilde{\vo}_n\sim\pit(\cdot\mid\vv,\vq^{\phi},\vp^{\phi})$ from the frozen self-teacher, given the video, the question, and the privileged information.
We retain the annotation only if all $N$ responses reproduce the pseudo-answer:
\begin{equation}
\gate(\vq^{\phi},\va^{\phi},\vp^{\phi})=
\prod_{n=1}^{N}
\ind[\ans(\tilde{\vo}_n)=\va^{\phi}],
\label{eq:gate}
\end{equation}
where $\ind[\cdot]$ is the indicator function and $\ans(\cdot)$ extracts the answer from a response.
The student is then trained only on the generated annotations, whose privileged information leads the teacher to the pseudo-answer (see \cref{sec:3.2}).

\subsection{Self-Distillation under a Joint Condition}
\label{sec:3.2}
In this section, we introduce how to train the student on the generated annotations.
From each retained annotation, we determine the \emph{direction} of each update by the verified pseudo-answer and the \emph{magnitude} of each token by the self-teacher.
In standard self-distillation, the teacher alone determines the direction of every update, without checking whether the rollout is correct.
As a result, tokens of correct rollouts can be penalized and tokens of incorrect rollouts can be encouraged (\cref{fig:motiv}-(b)).
Therefore, we let the pseudo-answer determine the direction, and use the self-teacher only to determine the magnitude of each token.
We then define the student loss under a joint condition, which updates a token only when the pseudo-answer and the self-teacher agree on the direction.

\paragraph{Direction from the pseudo-answer.}
The pseudo-answer determines the \textit{direction} of each rollout.
For each annotation retained by the Teacher-Answer gate, the student samples $G$ rollouts $\vo_i\sim\pis(\cdot\mid\vv,\vq^{\phi})$ from the video and the question.
Let $r_i\in\{0,1\}$ indicate whether the extracted answer of $\vo_i$ matches the pseudo-answer $\va^{\phi}$, and let $\bar{r}=\frac{1}{G}\sum_{i=1}^{G} r_i$ be the group match rate.
We assign each rollout a direction $s_i\in\{-1,0,1\}$:
\begin{equation}
s_i =
\begin{cases}
2r_i - 1 & \text{if } 0<\bar{r}<1,\\
0 & \text{otherwise.}
\end{cases}
\label{eq:sign}
\end{equation}
This direction equals the sign of the group-relative advantage $r_i-\bar{r}$ in GRPO~\cite{shao2024deepseekmath}.
Groups with $\bar{r}\in\{0,1\}$ receive $s_i=0$ for all $i$ and are excluded from distillation.
This motivates us to reward the MLLM annotator for questions that yield both correct and incorrect rollouts (\cref{sec:3.3}).

\paragraph{Magnitude from the self-teacher.}
The direction determines whether a token is encouraged or penalized, but not the magnitude of the update.
The self-teacher determines this \textit{magnitude} by evaluating each sampled token given the same video $\vv$, question $\vq^{\phi}$, and preceding tokens $\vo_{i,<t}$ as the student, together with the privileged information $\vp^{\phi}$.
We use the absolute value of the resulting log ratio $\tlr_{i,t}$ as the magnitude, where the log ratio decomposes as
\begin{equation}
\begin{aligned}
\tlr_{i,t}
={}&\log\frac{\pit(o_{i,t}\mid\cdot,\vp^{\phi},\vo_{i,<t})}{\pis(o_{i,t}\mid\cdot,\vo_{i,<t})}\\
={}&\underbrace{\log\frac{\pit(o_{i,t}\mid\cdot,\vp^{\phi},\vo_{i,<t})}{\pit(o_{i,t}\mid\cdot,\vo_{i,<t})}}_{\text{effect of privileged information}}
+\underbrace{\log\frac{\pit(o_{i,t}\mid\cdot,\vo_{i,<t})}{\pis(o_{i,t}\mid\cdot,\vo_{i,<t})}}_{\text{change from initialization}},
\end{aligned}
\label{eq:decomp}
\end{equation}
where $\cdot=(\vv,\vq^{\phi})$.
The first term measures how the privileged information changes the log-probability of token $o_{i,t}$ under the self-teacher, and serves as the learning signal.
Since the self-teacher is a frozen copy of the initial student, the second term measures how far the student has deviated from the initial student.
Under the joint condition, this term restrains further deviation, similar to the KL penalty toward the reference model in GRPO~\cite{shao2024deepseekmath,guo2025deepseek}.

\paragraph{Joint condition.}
The joint condition combines the \textit{direction} from the pseudo-answer with the \textit{magnitude} from the self-teacher.
A token is updated only when the log ratio $\tlr_{i,t}$ has the same sign as the direction $s_i$ of the rollout:
\begin{equation}
\rho_{i,t}=s_i\,\relu{s_i\,\tlr_{i,t}},\qquad
\rho_{i,t}\neq0\iff s_i\,\tlr_{i,t}>0,
\label{eq:agree}
\end{equation}
where $[\cdot]_{+}=\max(\cdot,0)$ and $\rho_{i,t}$ is the weight of token $o_{i,t}$ in the student loss.
A token that satisfies this condition receives the direction $s_i$ and the magnitude $|\tlr_{i,t}|$.
A token of a correct rollout is encouraged only when the self-teacher assigns the token a higher probability than the student, and a token of an incorrect rollout is penalized only in the opposite case.
As a result, the update never penalizes tokens of correct rollouts or encourages tokens of incorrect rollouts.

\paragraph{Student loss.}
We minimize the student loss $\mathcal{L}$, in which the distillation loss $\mathcal{L}_{\mathrm{dist}}$ weights each token by $\rho_{i,t}$:
\begin{equation}
\begin{aligned}
\mathcal{L}_{\mathrm{dist}}(\vq^{\phi})
&=-\frac1G\sum_{i=1}^{G}\frac{1}{|\vo_i|}
\sum_{t\notin\Uset_i}\sg(\rho_{i,t})\\
&\hspace{6.5em}\log\pSd,\\
\mathcal{L}
&=\frac{1}{|\Qset|}\sum_{\vq^{\phi}\in\Qset}
\big(\mathcal{L}_{\mathrm{dist}}(\vq^{\phi})+\lambda\mathcal{L}_{\mathrm{rep}}(\vq^{\phi})),
\end{aligned}
\label{eq:loss}
\end{equation}
where $\Qset$ contains the gated questions of a training step and $\lambda$ is the weight of the regularization loss $\mathcal{L}_{\mathrm{rep}}$.
The stop-gradient operator $\sg(\cdot)$ treats each token reward as a constant weight.
The regularization loss prevents the student from collapsing into repetitive outputs.
Following the sequence-level unlikelihood objective~\cite{welleck2019neuraltextgenerationunlikelihood}, it lowers the probability of the repeated tokens $\Uset_i$ in rollouts that loop.

\subsection{Learning an Informative MLLM Annotator}
\label{sec:3.3}

Only questions that yield both correct and incorrect rollouts contribute to distillation by \cref{eq:sign}.
Which questions satisfy this condition depends on the current student.
The student is updated at every training step.
However, a frozen MLLM annotator generates questions without feedback from these updates.
We therefore train the MLLM annotator online with a learnability reward computed from the rollouts sampled for student training (\cref{fig:motiv}-(c)).

\paragraph{Annotator reward.}
The learnability reward~\cite{rutherford2024no,foster2026learningreasonfrontierlearnability} is the variance of $r_i$ over the $G$ rollouts, scaled to $[0,1]$:
\begin{equation}
\Rann(\vq^{\phi})=\learn(\vq^{\phi})=\frac{4}{G}\sum_{i=1}^{G}(r_i-\rbar)^2=4\,\rbar\,(1-\rbar).
\label{eq:reward}
\end{equation}
This reward follows the analysis of reinforcement learning with binary rewards~\cite{bae2026online}.
The analysis lower-bounds the expected policy improvement on a question by the variance of the binary reward.
The reward is largest at $\rbar=1/2$, where equal numbers of rollouts receive $s_i=1$ and $s_i=-1$.
Questions with $\rbar\in\{0,1\}$, which \cref{eq:sign} excludes from distillation, receive zero reward.
Maximizing this reward trains the MLLM annotator to generate questions that yield both correct and incorrect rollouts for the current student.

\paragraph{Annotator advantage.}
We update the MLLM annotator with GRPO over the $M$ candidate annotations of each video, alongside the student update.
For each video, let $\Cset$ be the indices of the candidates retained by the Teacher-Answer gate.
The advantage of candidate $j$ is
\begin{equation}
\Adv_j=
\begin{cases}
\dfrac{\learn(\vq^{\phi}_j)-\mu_{\Cset}}{\sigma_{\Cset}+\epsilon}, & j\in\Cset,\\[1.2ex]
0, & \text{rejected by the gate},\\[0.3ex]
-1, & \text{invalid annotation},
\end{cases}
\label{eq:adv}
\end{equation}
where $\mu_{\Cset}$ and $\sigma_{\Cset}$ are the mean and standard deviation of the learnability rewards over $\Cset$, and $\epsilon$ is a small constant for numerical stability.
Candidates rejected by the Teacher-Answer gate receive zero advantage instead of a penalty, so that the MLLM annotator does not avoid question types with a high rejection rate.
The negative advantage for candidates with an invalid annotation trains the MLLM annotator to follow the annotation format.

%% file: sec/4_Experiment.tex
\section{Experiment}
\label{sec:4}

% =============================================================================
% ================================ Qwen3-VL-4B ================================
% =============================================================================

\begin{table*}[th]
\centering
\caption{\textbf{Results with Qwen3-VL-4B and Qwen3-VL-8B.} Small numbers next to our scores show the change from the Qwen3-VL model of the same size. Annot.-free marks methods that train without human-annotated questions, answers, or privileged information. Best results within each model size are in \textbf{bold}. V-MME: Video-MME, MVB: MVBench, LVB: LongVideoBench, TempC: TempCompass, VSI: VSI-Bench, Holmes: Video-Holmes, WS: WorldSense, QBench: QBench-Video.}
\vspace{-0.5em}
\label{tab:1}
\setlength{\tabcolsep}{4pt}
\setlength{\aboverulesep}{0pt}
\setlength{\belowrulesep}{0pt}
\renewcommand{\arraystretch}{1.25}
\resizebox{\textwidth}{!}{%
\begin{tabular}{lc*{11}{l}}
\toprule
Method & Annot.-free & MVB & MLVU & LVB & V-MME & V-MME-v2 & TempC & VSI & Holmes & WS & QBench & Avg. \\
\midrule
Qwen3-VL-4B & -- & 62.0 & 53.4 & 51.3 & 57.9 & 24.4 & 71.1 & 45.7 & 34.8 & 36.3 & 44.7 & 48.2 \\
Video-OPSD (4B) & \xmark & 62.4 & 56.0 & 53.0 & 58.7 & 24.5 & 70.9 & 47.0 & 36.4 & 38.3 & 45.7 & 49.3 \\
Video-Zero (4B) & \cmark & 62.7 & 53.3 & 51.8 & 59.7 & 25.2 & 70.6 & 45.8 & 36.1 & 37.0 & 44.5 & 48.7 \\
\rowcolor{ourspurple}
\textbf{Ours (4B)} & \cmark & \textbf{63.9}\,\gainup{1.9} & \textbf{62.6}\,\gainup{9.2} & \textbf{57.4}\,\gainup{6.1} & \textbf{59.8}\,\gainup{1.9} & \textbf{26.2}\,\gainup{1.8} & \textbf{75.1}\,\gainup{4.0} & \textbf{47.4}\,\gainup{1.7} & \textbf{38.9}\,\gainup{4.1} & \textbf{40.6}\,\gainup{4.3} & \textbf{47.3}\,\gainup{2.6} & \textbf{51.9}\,\gainup{3.7} \\
\midrule
Qwen3-VL-8B & -- & 66.0 & 59.4 & 56.0 & 60.4 & 25.7 & \textbf{79.1} & 43.7 & 39.6 & 39.7 & 46.1 & 51.6 \\
Video-OPSD (8B) & \xmark & 66.0 & 58.1 & 56.5 & 60.0 & 26.0 & 78.9 & 44.1 & 37.5 & 40.0 & 46.8 & 51.4 \\
Video-Zero (8B) & \cmark & 65.1 & 59.3 & 55.9 & 61.1 & 26.6 & 78.5 & 44.5 & 39.7 & 38.8 & \textbf{47.3} & 51.7 \\
\rowcolor{ourspurple}
\textbf{Ours (8B)} & \cmark & \textbf{66.5}\,\gainup{0.5} & \textbf{62.6}\,\gainup{3.2} & \textbf{57.9}\,\gainup{1.9} & \textbf{61.4}\,\gainup{1.0} & \textbf{27.7}\,\gainup{2.0} & 78.8\,\gaindn{0.3} & \textbf{47.8}\,\gainup{4.1} & \textbf{40.9}\,\gainup{1.3} & \textbf{42.2}\,\gainup{2.5} & 46.0\,\gaindn{0.1} & \textbf{53.2}\,\gainup{1.6} \\
\bottomrule
\end{tabular}}
\vspace{-1em}
\end{table*}

% =============================================================================
% =============================================================================
% =============================================================================

% --------------------------------------------------------------------------------
% ---------------- GT privileged evidence vs online annotation ----------------
% --------------------------------------------------------------------------------

\begin{table}[th]
\centering
\caption{\textbf{Ablation study on annotation sources.} Human Annot.\ replaces the MLLM annotator in our method with human-annotated questions, answers, and privileged information from Video-Holmes and STAR. Avg.\ is the average over the eight benchmarks of \cref{tab:1} outside the training domain. Human Annot.\ leads in the training domain, while our method achieves a higher average outside the training domain. $\Delta$: Ours $-$ Human Annot.}
\vspace{-0.5em}
\label{tab:2}
\setlength{\tabcolsep}{4pt}
\setlength{\aboverulesep}{0pt}
\setlength{\belowrulesep}{0pt}
\renewcommand{\arraystretch}{1.25}
\scalebox{0.85}{
\begin{tabular}{lcccc}
\toprule
Benchmark & \textcolor{gray}{Base} & Human Annot.\ & \textbf{Ours} & $\Delta$ \\
\midrule
\rowcolor{indomain}
\textit{In training domain} & & & & \\
Video-Holmes & \textcolor{gray}{34.8} & \textbf{49.4} & 38.9 & $-$10.5 \\
MVBench & \textcolor{gray}{62.0} & \textbf{66.0} & 63.9 & $-$2.1 \\
\midrule
\rowcolor{outdomain}
\textit{Out of training domain} & & & & \\
QBench-Video & \textcolor{gray}{44.7} & 40.8 & \textbf{47.3} & +6.5 \\
MLVU & \textcolor{gray}{53.4} & 61.3 & \textbf{62.6} & +1.3 \\
Video-MME-v2 & \textcolor{gray}{24.4} & 25.3 & \textbf{26.2} & +0.9 \\
LongVideoBench & \textcolor{gray}{51.3} & 56.2 & \textbf{57.4} & +1.2 \\
\cmidrule(lr){1-5}
Avg.\ (8 benchmarks) & \textcolor{gray}{48.1} & 51.1 & \textbf{52.0} & +0.9 \\
\bottomrule
\end{tabular}}
\vspace{-0.5em}
\end{table}

% =============================================================================
% =============================================================================
% =============================================================================

% --------------------------------------------------------------------------------
% ---------------------- frozen annoator vs online annotation --------------------
% --------------------------------------------------------------------------------

\begin{table}[th]
\centering
\caption{\textbf{Ablation study on annotator training.} \emph{Frozen} keeps the MLLM annotator fixed during training, while all other settings follow our method. \emph{Online} is our method, which trains the MLLM annotator with the reward from the current student. Avg.\ is the average over the 10 benchmarks. $\Delta$: Online $-$ Frozen.}
\vspace{-0.5em}
\label{tab:3}
\setlength{\tabcolsep}{6pt}
\setlength{\aboverulesep}{0pt}
\setlength{\belowrulesep}{0pt}
\renewcommand{\arraystretch}{1.3}
\scalebox{0.85}{
\begin{tabular}{lcccc}
\toprule
\multirow{2}{*}{Benchmark} & \multirow{2}{*}{\textcolor{gray}{Base}} & \multicolumn{2}{c}{MLLM annotator} & \multirow{2}{*}{$\Delta$} \\
\cmidrule(lr){3-4}
 & & Frozen & \textbf{Online (Ours)} & \\
\midrule
Avg.\ & \textcolor{gray}{48.2} & 51.3 & \textbf{51.9} & +0.6 \\
\bottomrule
\end{tabular}}
\vspace{-1em}
\end{table}

% =============================================================================
% =============================================================================
% =============================================================================

\subsection{Setup}

\paragraph{Training and evaluation details.}
We collect only raw videos from the official training splits of STAR~\cite{wu2024star}, NExT-QA~\cite{xiao2021next}, Perception Test~\cite{patraucean2023perception}, and Video-Holmes~\cite{cheng2025video}, and the MLLM annotator generates all questions, pseudo-answers, and privileged information online.
For Video-Holmes, all training data, including the human annotations used in \cref{tab:2}, are taken from the training split, and all results are measured on the test split.
No training video appears in the evaluation benchmarks, including MVBench, which partly uses videos from STAR and Perception Test.

We evaluate on 10 video benchmarks, Video-MME~\cite{fu2025video}, Video-MME-v2~\cite{fu2026video}, MVBench~\cite{li2024mvbench}, MLVU~\cite{zhou2025mlvu}, Long\-Video\-Bench~\cite{wu2024longvideobench}, Temp\-Compass~\cite{liu2024tempcompass}, VSI-Bench~\cite{yang2025thinking}, Video-Holmes~\cite{cheng2025video}, WorldSense~\cite{hong2026worldsense}, and QBench-Video~\cite{zhang2024q}, which cover general video understanding, temporal perception, spatial reasoning, complex reasoning, real-world understanding.
We use only questions scored by answer matching, so that no score depends on an external judge model.
For MLVU, TempCompass, and QBench-Video, we therefore use only the multiple-choice subsets.
For WorldSense, all models receive only the video frames without the audio track.
For a fair comparison, we evaluate all models with VLMEvalKit~\cite{duan2024vlmevalkit}, a widely used open-source evaluation toolkit.
All models, including the baselines, share the same prompt, answer extraction, and 32 uniformly sampled frames as VLMEvalKit, and we report the official metric of each benchmark.
Absolute scores can differ from those under model-specific settings, such as more frames or different prompts, but all comparisons in this paper use identical settings.
Video-OPSD and all ablation variants are trained with the same prompt and answer format as our method.
Any gain from format adaptation is thus shared among these models, and the differences among them reflect the training signal.

\paragraph{Baselines.}
We compare against the base model and two methods that post-train it with self-generated or privileged supervision.
Video-OPSD~\cite{wang2026video} has no public implementation, so we reimplement it following the paper.
Because Video-OPSD requires privileged-information annotations, we train it on the two datasets that provide those annotations:
Video-Holmes~\cite{cheng2025video} and STAR~\cite{wu2024star}.
For Video-Zero~\cite{zhang2026video}, we use the official released checkpoints.

\paragraph{Implementation details.}
We use Qwen3-VL-4B-Instruct and Qwen3-VL-8B-Instruct~\cite{bai2025qwen3} as the student, and the frozen initial student as the self-teacher.
The MLLM annotator is the same model as the student, with LoRA~\cite{hu2021lora}, and is trained with GRPO.
We train for 50 steps with 16 videos per step.
We use $M=4$ candidate annotations, $N=4$ self-teacher responses, and $G=8$ rollouts.
The learning rate of the student decays linearly from $1\times10^{-5}$ to $1\times10^{-7}$, and that of the MLLM annotator decays from $5\times10^{-6}$ to 0.

\subsection{Analysis}

\subsubsection{Main results}
\Cref{tab:1} reports the results on 10 video benchmarks with Qwen3-VL-4B and Qwen3-VL-8B.
Our method achieves the best average with both model sizes, without any human annotation.
Specifically, our method improves all 10 benchmarks over Qwen3-VL-4B and raises the average from 48.2 to 51.9.
Our method improves Qwen3-VL-4B on all 10 benchmarks, which cover diverse tasks such as temporal perception, spatial reasoning, and visual quality assessment.
These results suggest that our method improves the student in a task-agnostic manner, since the method uses neither task-specific rewards nor task-specific annotations.

Our method also improves Qwen3-VL-8B.
The average increases from 51.6 to 53.2, with the best results on 8 of the 10 benchmarks.
In contrast, Video-Zero and Video-OPSD change the average by at most 0.2 points.
This gap suggests that informative questions and token-level supervision together continue to improve the student even when the base model already performs well.
These results show that the gains of our method are not limited to a specific model size.

Our method outperforms Video-OPSD, although Video-OPSD trains on human-annotated questions and privileged information.
Our method requires only raw videos, yet achieves a higher average by 2.6 points with the 4B model and 1.8 points with the 8B model.
The gain appears on all 10 benchmarks with the 4B model and on 8 benchmarks with the 8B model.
This result indicates that the MLLM annotator can replace human annotations for OPSD on video.
Our method also outperforms Video-Zero, which trains without annotations as well, by 3.2 points with the 4B model and 1.5 points with the 8B model.
Both methods generate questions from raw videos, but Video-Zero trains the student with GRPO, which assigns a single reward to each rollout.
In contrast, our method trains the student with OPSD, where the self-teacher provides token-level supervision on each rollout.
This gap suggests that OPSD is more suitable than GRPO for annotation-free training.

The gains of our method do not result from correcting the answer format.
The base model, Qwen3-VL-4B, produces answers in various invalid formats, which are scored as incorrect.
We therefore evaluate both models only on the questions that the base model answers in a valid format.
This setting favors the base model, since the exclusion removes only incorrect answers of this model but can also remove correct answers of our method.
Even in this setting, our method achieves an average accuracy of 53.0, while the base model achieves 51.7.
Our method thus improves the answers themselves.

\subsubsection{Does annotation-free training generalize beyond annotated domains?}
As shown in \cref{tab:2}, our method outperforms a variant trained on human annotations on benchmarks outside the annotated domains.
This variant replaces the MLLM annotator with human annotations from Video-Holmes and STAR.
Video-Holmes and MVBench, which partly uses videos from STAR, are thus in the training domain.
The variant achieves higher accuracy on these benchmarks, since the variant is trained on human-annotated questions from the training domain.
The verifiable reward also allows the variant to outperform Video-OPSD, which trains on the same annotations.
However, our method achieves an average of 52.0 outside the training domain, 0.9 points higher than the variant.
The variant depends on datasets with human-annotated privileged information, which cover only a few domains.
In contrast, our method requires only raw videos, so the student trains on more datasets, and the gains extend beyond the annotated domains.
Overall, our method enables task-agnostic post-training that generalizes across video domains without human annotations.

\begin{table}[t]
\centering
\caption{\textbf{Additional ablation studies.} We compare our method with a variant without data augmentation (DA), which trains only on raw videos from Video-Holmes and STAR instead of all four datasets. Another variant is trained without the Teacher-Answer (TA) gate. We also compare with a variant that replaces OPSD with GRPO on the pseudo-answers. The last variant uses the KL objective of standard OPSD instead of the joint condition, so the rollouts are not verified against the pseudo-answers and the self-teacher alone determines the direction of every update.
}
\vspace{-0.5em}
\label{tab:4}
\setlength{\tabcolsep}{9pt}
\setlength{\aboverulesep}{0pt}
\setlength{\belowrulesep}{0pt}
\renewcommand{\arraystretch}{1.3}
\scalebox{0.85}{
\begin{tabular}{ccccc}
\toprule
DA & TA Gate & OPSD & Verifiable Reward & Avg. \\
\midrule
\rowcolor{indomain}
\xmark & \cmark & \cmark & \cmark & 50.8 \\
\rowcolor{indomain}
\cmark & \xmark & \cmark & \cmark & 51.0 \\
\rowcolor{indomain}
\cmark & \cmark & \xmark & \xmark & 48.5 \\
\rowcolor{indomain}
\cmark & \cmark & \cmark & \xmark & 50.8 \\
\rowcolor{ourspurple}
\cmark & \cmark & \cmark & \cmark & \textbf{51.9} \\
\bottomrule
\end{tabular}}
\vspace{-1em}
\end{table}

\subsubsection{Does online training of the MLLM annotator make the annotations more informative?}
We compare our method with a variant that keeps the MLLM annotator frozen during training.
Both settings generate annotations at every training step and differ only in whether the MLLM annotator is updated with the reward from the current student.
\Cref{fig:motiv}-(c) shows how the informative questions change during training.
The frozen annotator generates 122 questions that are informative for the initial student.
For the student after 50 training steps, only 52 of these questions remain informative.
Whether a question is informative thus depends on the current student, and many questions generated for the initial student lose this property as the student learns.
The frozen annotator receives no feedback from the student, and the percentage of informative questions from this annotator decreases from 48.2\% in the first block to 28.9\% in the last block.
In contrast, the online annotator is rewarded for questions on which the current student produces both correct and incorrect rollouts (\cref{sec:3.3}).
In the last block, the percentage for the online annotator is 32.6\%, 3.7 points higher than that for the frozen annotator.
Online training thus makes the MLLM annotator generate more informative questions for the current student.

\Cref{tab:3} shows that these more informative questions further improve the student.
The frozen annotator already improves the average of Qwen3-VL-4B from 48.2 to 51.3, which shows that the generated annotations provide useful supervision for OPSD.
Online training further improves the average to 51.9.
Since only annotator training differs, this gain results from the annotations generated by the online annotator.
These results support our motivation that the training data should change as the student evolves.

\subsection{Ablation study}
\label{sec:ablation}
\Cref{tab:4} ablates data augmentation, the Teacher-Answer gate, OPSD, and the direction from the pseudo-answer.
Each component contributes to the performance, and our method with all components achieves the highest average of 51.9.
Data augmentation raises the average from 50.8 to 51.9.
Since the MLLM annotator requires no human annotations, the student can be trained on raw videos from more datasets, which improves performance across diverse tasks.
The Teacher-Answer gate improves the average from 51.0 to 51.9.
The gate filters out annotations whose pseudo-answers the self-teacher does not reproduce, which can include wrong pseudo-answers.
OPSD yields the largest gain, from 48.5 to 51.9.
GRPO on the pseudo-answers assigns a single reward to each rollout, whereas OPSD provides token-level supervision from the self-teacher.
The direction from the pseudo-answer gives a further gain of 1.1 points over standard OPSD.
In standard OPSD, the self-teacher alone determines the direction of every update, so the student can be encouraged on tokens of incorrect rollouts and penalized on tokens of correct ones.
Verifying each rollout against the pseudo-answer prevents such updates.

%% file: sec/5_Conclusion.tex
\section{Conclusion}
\label{sec:5}
We proposed MLLM-as-an-Annotator, an online data generation loop in which an MLLM annotator replaces the human annotations required for OPSD on video.
Without any human annotation, our method improves Qwen3-VL-4B by 3.7 points on average across 10 video benchmarks and outperforms OPSD trained on human annotations.
These results suggest that the annotations for video post-training can be learned during training rather than collected in advance.

%% file: sec/X_suppl.tex
\clearpage
\setcounter{page}{1}
\maketitlesupplementary
\raggedbottom

\section{Analysis of the Learnability Reward}
\label{sec:app_reward}

This section justifies the annotator reward in \cref{eq:reward}.
Let $p\in[0,1]$ denote the probability that a rollout of the current student matches the pseudo-answer $\va^{\phi}$ of a question $\vq^{\phi}$.
The $G$ rollouts are sampled independently, so $r_1,\dots,r_G$ are i.i.d.\ Bernoulli$(p)$ variables, and $\rbar=\frac{1}{G}\sum_{i=1}^{G}r_i$.

\paragraph{Variance of binary rewards.}
The learnability reward equals the scaled variance of the binary rewards.
This variance is largest when half of the rollouts are correct.
\begin{proposition}[Variance of binary rewards]
\label{prop:var}
For $r_i\in\{0,1\}$, $\frac{1}{G}\sum_{i=1}^{G}(r_i-\rbar)^2=\rbar(1-\rbar)$.
Hence $\learn(\vq^{\phi})=4\rbar(1-\rbar)\in[0,1]$, with the maximum $1$ at $\rbar=1/2$ for even $G$ and the minimum $0$ at $\rbar\in\{0,1\}$.
\tcblower
\begin{proof}
Since $r_i\in\{0,1\}$, $r_i^2=r_i$. Then
\begin{equation*}
\frac{1}{G}\sum_{i=1}^{G}(r_i-\rbar)^2
=\frac{1}{G}\sum_{i=1}^{G}r_i^2-\rbar^2
=\rbar(1-\rbar).
\end{equation*}
The function $x\mapsto4x(1-x)$ is concave on $[0,1]$, equals $1$ at $x=1/2$, and equals $0$ at $x\in\{0,1\}$.
The value $\rbar=1/2$ is attained when $G$ is even, which holds for $G=8$.
\end{proof}
\end{proposition}

\paragraph{Expected reward.}
The expected reward is proportional to the variance of the binary reward of the student, as stated in \cref{prop:exp}.
Questions with $p$ near $0$ or $1$ receive a reward near $0$.
Maximizing the reward therefore favors questions with $p$ close to $1/2$ for the current student.
Under a KL-regularized objective, Bae~\etal~\cite{bae2026online} show that the divergence between the initial and the optimal policy on a question is lower-bounded by $p(1-p)$ up to a constant, to second order, which motivates the reward.
\begin{proposition}[Expected reward]
\label{prop:exp}
$\mathbb{E}\big[\learn(\vq^{\phi})\big]=\big(1-\tfrac{1}{G}\big)\,4p(1-p)$.
\tcblower
\begin{proof}
$\mathbb{E}[\rbar]=p$ and $\mathrm{Var}[\rbar]=p(1-p)/G$, so $\mathbb{E}[\rbar^2]=p^2+p(1-p)/G$.
Then
\begin{equation*}
\begin{aligned}
\mathbb{E}\big[4\rbar(1-\rbar)\big]
&=4\Big(p-p^2-\frac{p(1-p)}{G}\Big)\\
&=\Big(1-\frac{1}{G}\Big)\,4p(1-p).
\end{aligned}
\end{equation*}
\end{proof}
\end{proposition}

\paragraph{Questions that contribute to distillation.}
By \cref{eq:sign}, a question contributes to distillation only if $0<\rbar<1$, that is, only if the rollouts include both correct and incorrect ones.
\begin{proposition}[Contribution to distillation]
\label{prop:mixed}
The probability $q(p)=1-p^{G}-(1-p)^{G}$ that a question contributes to distillation is zero at $p\in\{0,1\}$, increases as $p$ approaches $1/2$, and is maximized at $p=1/2$ for $G\ge2$.
\tcblower
\begin{proof}
$q(0)=q(1)=0$, and $q(p)=q(1-p)$.
The derivative is $q'(p)=G\big[(1-p)^{G-1}-p^{G-1}\big]$, which is positive for $p<1/2$ and zero only at $p=1/2$.
The second derivative $q''(p)=-G(G-1)\big[p^{G-2}+(1-p)^{G-2}\big]$ is negative for $G\ge2$, so $p=1/2$ is the maximum.
\end{proof}
\end{proposition}
The expected reward in \cref{prop:exp} and $q(p)$ both increase as $p$ approaches $1/2$.
A question with a higher expected reward thus contributes to distillation with a higher probability.
Moreover, any question with $\learn(\vq^{\phi})>0$ satisfies $0<\rbar<1$ and thus contributes to distillation by \cref{eq:sign}.

\begin{table*}[t]
\centering
\caption{\textbf{Training on videos from Open-o3 Video.} \emph{Ours (Open-o3)} trains our method on 600 raw videos sampled from Open-o3 Video~\cite{meng2025open}, the video source of Video-Zero~\cite{zhang2026video}. All models start from Qwen3-VL-4B~\cite{bai2025qwen3}. The benchmark names follow \cref{tab:1}.}
\label{tab:app_o3}
\setlength{\tabcolsep}{7pt} % <- \small에서 본문 폭에 맞춘 값
\setlength{\aboverulesep}{0pt}
\setlength{\belowrulesep}{0pt}
\renewcommand{\arraystretch}{1.25}
\small
\begin{tabular}{lccccccccccc}
\toprule
Method & MVB & MLVU & LVB & V-MME & V-MME-v2 & TempC & VSI & Holmes & WS & QBench & Avg. \\
\midrule
Qwen3-VL-4B     & 62.0 & 53.4 & 51.3 & 57.9 & 24.4 & 71.1 & 45.7 & 34.8 & 36.3 & \textbf{44.7} & 48.2 \\
Video-Zero (4B) & 62.7 & 53.3 & 51.8 & \textbf{59.7} & 25.2 & 70.6 & 45.8 & 36.1 & 37.0 & 44.5 & 48.7 \\
\rowcolor{ourspurple}
\textbf{Ours (Open-o3)} & \textbf{65.1} & \textbf{60.3} & \textbf{57.1} & \textbf{59.7} & \textbf{26.2} & \textbf{75.7} & \textbf{46.1} & \textbf{39.5} & \textbf{40.6} & 43.4 & \textbf{51.4} \\
\bottomrule
\end{tabular}
\end{table*}

\paragraph{Balance between encouraged and penalized rollouts.}
For a question with $k$ correct rollouts, $\rbar=k/G$, and the sum of absolute deviations is
\begin{equation*}
\sum_{i=1}^{G}|r_i-\rbar|
=k(1-\rbar)+(G-k)\,\rbar
=2G\,\rbar(1-\rbar)
=\frac{G}{2}\,\learn(\vq^{\phi}).
\end{equation*}
This sum is thus proportional to the learnability reward and reaches the maximum at $\rbar=1/2$, where $k=G/2$ rollouts receive $s_i=1$ and the other $G/2$ rollouts receive $s_i=-1$.
The learnability reward is therefore largest for questions that yield the same number of correct rollouts to encourage and incorrect rollouts to penalize.

% ============================================================
\section{Effect of the Answer Format}
\label{sec:app_format}

\begin{table}[t]
\centering
\caption{\textbf{Accuracy by the validity of the answer format.} \emph{Invalid} is the percentage of responses from which no valid answer is extracted. \emph{All} counts invalid responses as incorrect, as in \cref{tab:1}. \emph{Base-valid} uses only the questions for which the base model gives a valid answer, and \emph{All-valid} uses only the questions for which all four models give a valid answer. For the base model, we report the accuracy. For the other models, we report the difference from the base model. All values are averaged over the 10 benchmarks, and all models are evaluated with the same seed and the same answer extraction.}
\label{tab:app_format}
\setlength{\tabcolsep}{6pt}
\setlength{\aboverulesep}{0pt}
\setlength{\belowrulesep}{0pt}
\renewcommand{\arraystretch}{1.25}
\resizebox{\linewidth}{!}{%
\begin{tabular}{lcccc}
\toprule
Method & Invalid (\%) & All & Base-valid & All-valid \\
\midrule
Qwen3-VL-4B     & 6.17 & 48.16 & 51.65 & 51.91 \\
Video-OPSD (4B) & 3.91 & $+$1.14 & $-$0.46 & $+$0.19 \\
Video-Zero (4B) & 5.67 & $+$0.52 & $-$0.87 & $+$0.28 \\
\rowcolor{ourspurple}
\textbf{Ours (4B)} & \textbf{0.75} & $\mathbf{+3.75}$ & $\mathbf{+1.35}$ & $\mathbf{+1.22}$ \\
\bottomrule
\end{tabular}}
\end{table}

This section examines whether the gain of our method remains when the answer format is controlled.
A response is invalid when no answer can be extracted from it, and an invalid response is scored as incorrect.
\Cref{tab:app_format} reports the accuracy on all questions and on the questions with valid answers.

Our method remains better than the base model on the questions with valid answers.
The gain is 1.35 points on the questions for which the base model gives a valid answer, and 1.22 points on the questions for which all models give a valid answer.
The gain is therefore not explained by the answer format alone.
In contrast, Video-OPSD~\cite{wang2026video} and Video-Zero~\cite{zhang2026video} do not exceed the base model on the questions for which the base model gives a valid answer.

Our method also produces valid answers more reliably.
The base model gives an invalid answer for 6.17\% of the questions, whereas our method does so for 0.75\%.
Invalid answers of the base model consist of answers that are not one of the options (2.42\%), missing or unclosed answer blocks (2.24\%), and responses truncated at the token limit (1.50\%), most of which are repetitive.
Our method produces no truncated response.

% ============================================================
\section{Training on the Video Source of Video-Zero}
\label{sec:app_o3}

To compare with Video-Zero on the same video source, we train our method on the videos that Video-Zero uses.
Video-Zero samples 600 raw videos from Open-o3 Video~\cite{meng2025open}.
The list of these videos is not released, so we randomly sample 600 raw videos from the same source and exclude the videos that appear in the evaluation benchmarks.
We use none of the questions or answers of Open-o3 Video.
All other settings follow \cref{sec:4}.
We compare with the fourth-iteration checkpoint of Video-Zero, which the authors recommend.

\Cref{tab:app_o3} shows that our method outperforms Video-Zero when both are trained on videos from the same source.
Our method reaches an average of 51.4, 2.7 points higher than Video-Zero, and is higher on 8 of the 10 benchmarks and equal on one.
The gain of our method over Video-Zero therefore does not depend on the choice of training videos.
% ============================================================
\section{Results on Another Model Family}
\label{sec:app_molmo}

\begin{table}[t]
\centering
\caption{\textbf{Results across architectures.} We apply our method to Qwen3-VL-4B and Molmo2-4B~\cite{clark2026molmo2} with the same training settings. Avg.\ is the average over the 10 benchmarks in \cref{tab:1}.}
\label{tab:5}
\setlength{\aboverulesep}{0pt}
\setlength{\belowrulesep}{0pt}
\renewcommand{\arraystretch}{1.2}
\scalebox{0.9}{%
\begin{tabular*}{1.0\columnwidth}{@{\extracolsep{\fill}}lccc}
\toprule
Architecture & Base & \textbf{Ours} & $\Delta$ \\
\midrule
Qwen3-VL-4B & 48.2 & \cellcolor{ourspurple}\textbf{51.9} & +3.7 \\
Molmo2-4B   & 50.6 & \cellcolor{ourspurple}\textbf{51.6} & +1.0 \\
\bottomrule
\end{tabular*}}
\vspace{-0.5em}
\end{table}

This section applies our method to a model family other than Qwen3-VL.
We use Molmo2-4B~\cite{clark2026molmo2}, which differs from Qwen3-VL-4B in the architecture and the training data.
The MLLM annotator, the student, and the self-teacher all start from Molmo2-4B.
All other settings follow \cref{sec:4}.

\Cref{tab:5} shows that our method improves both model families.
On Molmo2-4B, our method achieves 51.6, 1.0 point higher than the base model.
This gain holds even though the base Molmo2-4B already scores 2.4 points higher than the base Qwen3-VL-4B.
Our method therefore does not depend on a specific model family.

% ============================================================
\section{Qualitative Results}
\label{sec:app_qual}

\subsection{Annotations Retained and Rejected by the Teacher-Answer Gate}
\label{sec:app_qual_gate}

\begin{figure*}[th!]
  \centering
  \includegraphics[width=0.8\linewidth]{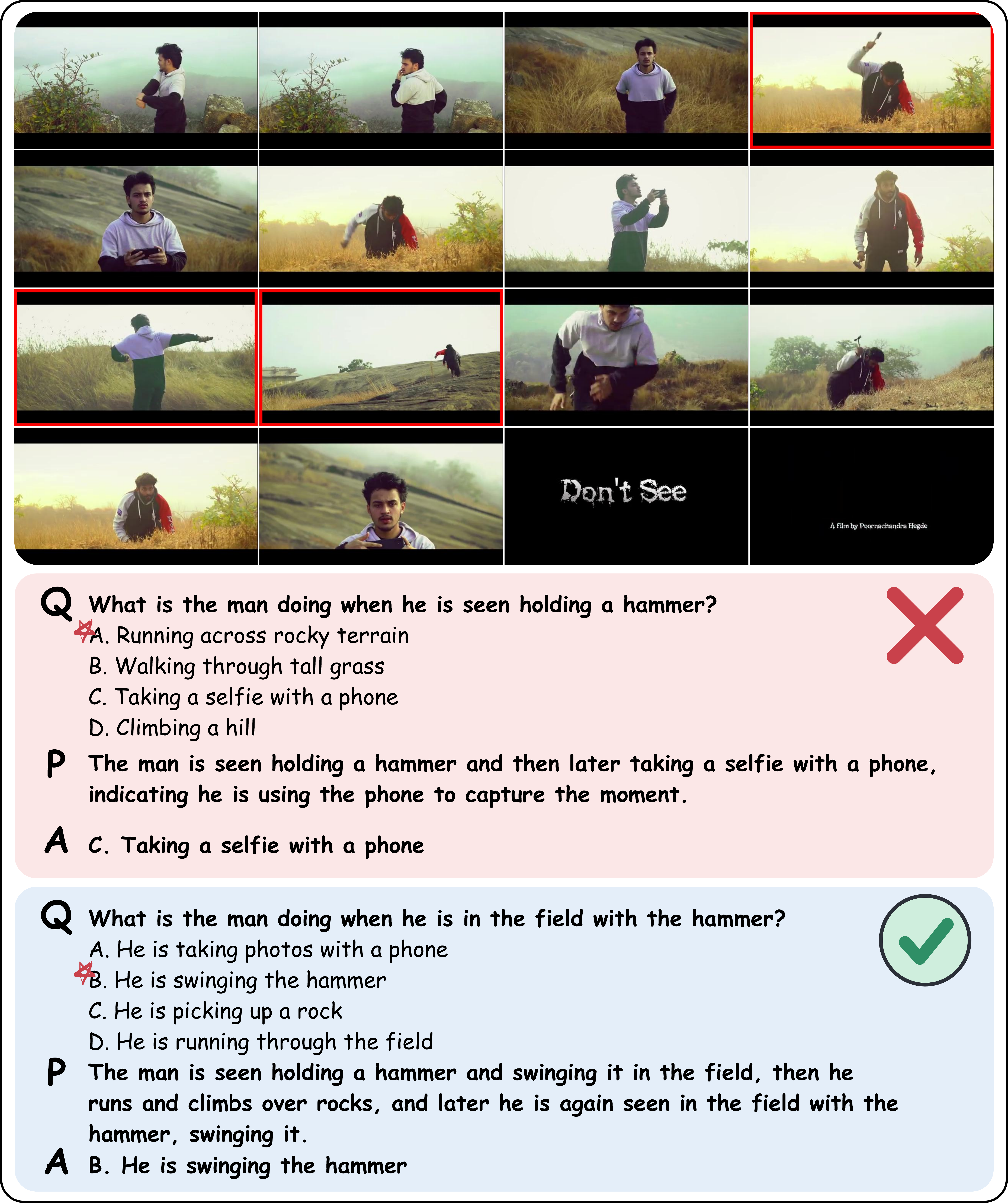}
  \caption{\textbf{Annotations rejected (top) and retained (bottom) by the Teacher-Answer gate on Video-Holmes.}}
  \label{fig:qual1_1}
  \vspace{-1em}
\end{figure*}

\begin{figure*}[th!]
  \centering
  \includegraphics[width=0.8\linewidth]{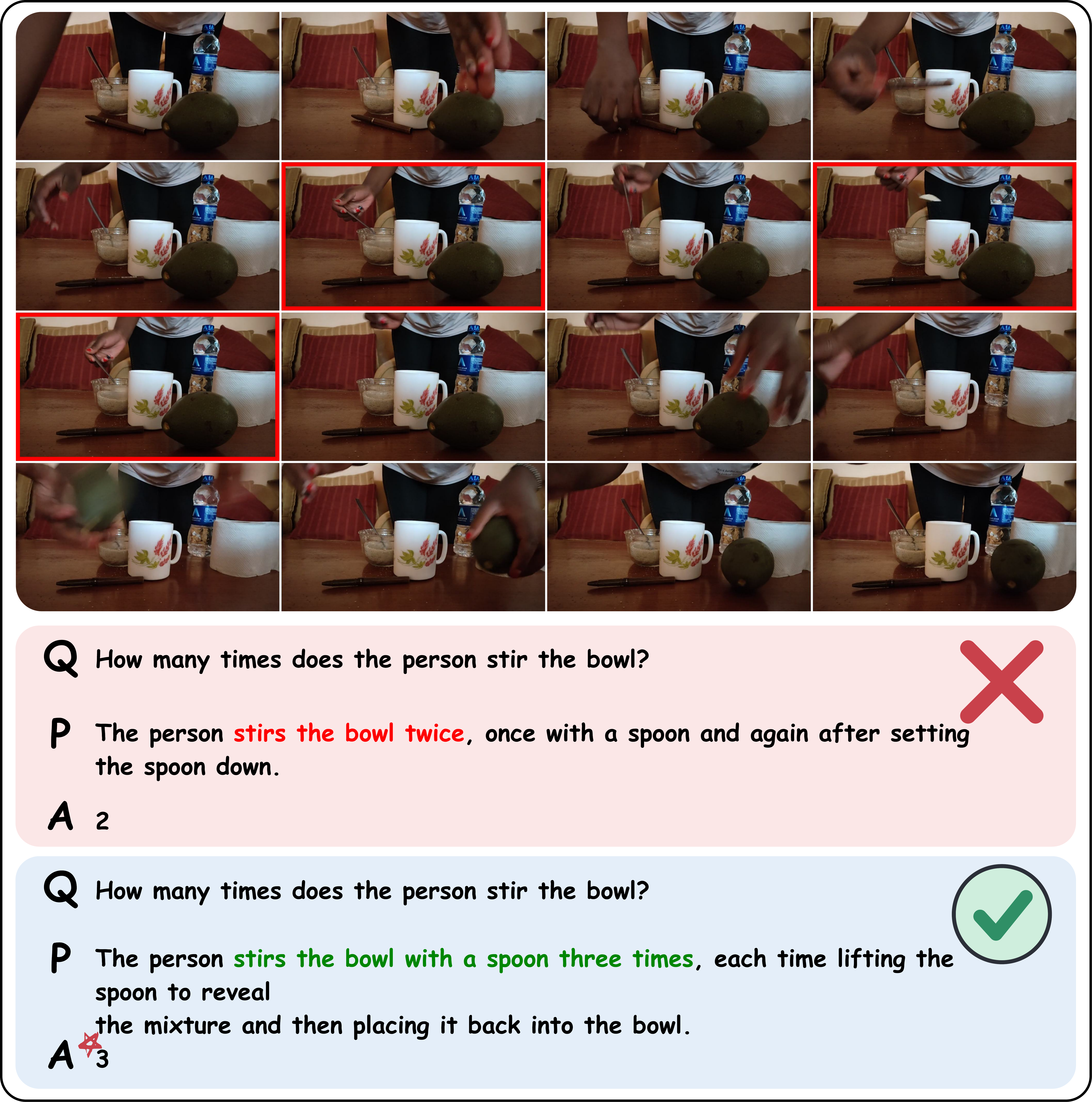}
  \caption{\textbf{Annotations rejected (top) and retained (bottom) by the Teacher-Answer gate on Perception Test.}}
  \label{fig:qual1_2}
  \vspace{-1em}
\end{figure*}

\begin{figure*}[th!]
  \centering
  \includegraphics[width=0.8\linewidth]{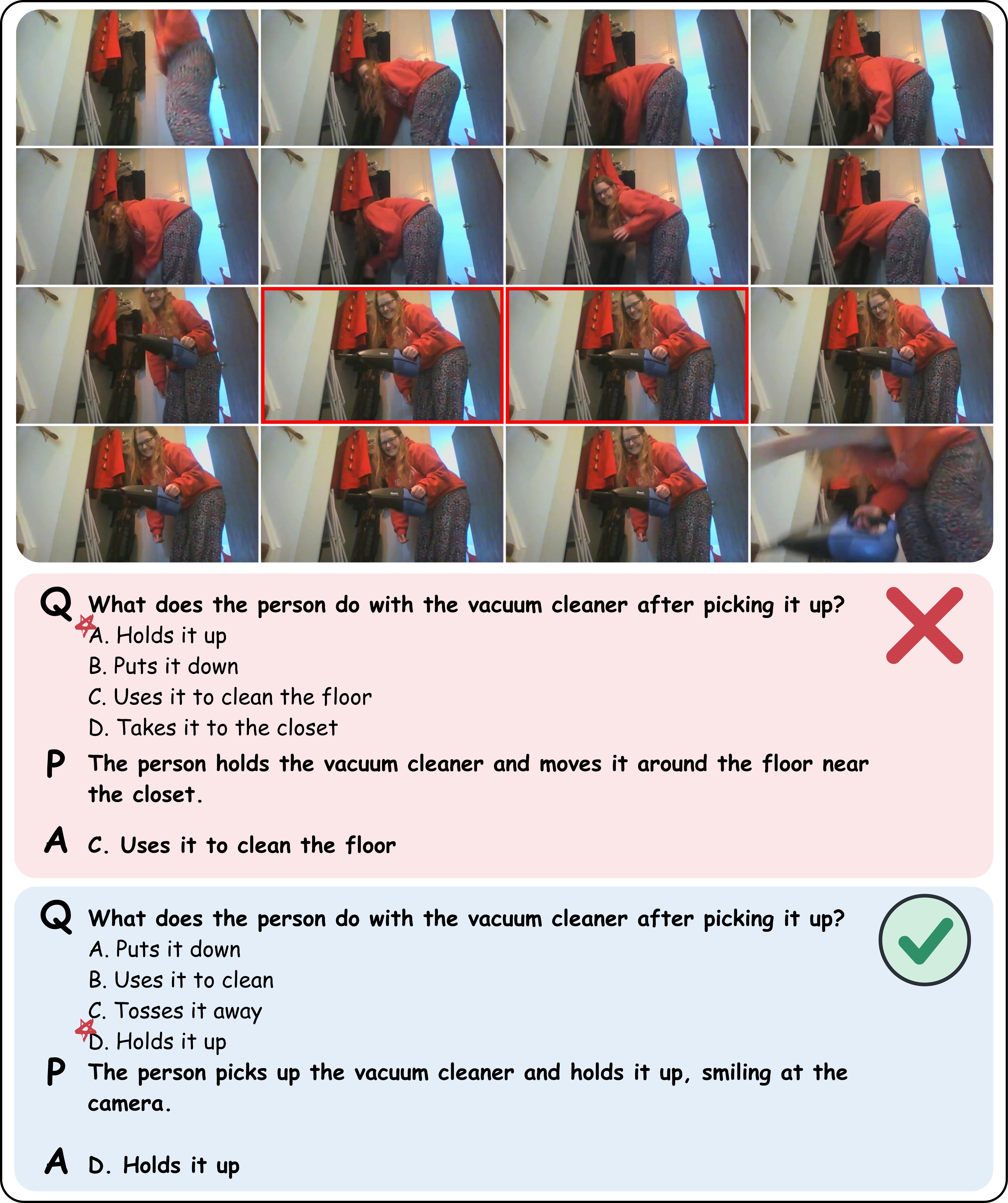}
  \caption{\textbf{Annotations rejected (top) and retained (bottom) by the Teacher-Answer gate on STAR.}}
  \label{fig:qual1_3}
  \vspace{-1em}
\end{figure*}

\begin{figure*}[th!]
  \centering
  \includegraphics[width=0.7\linewidth]{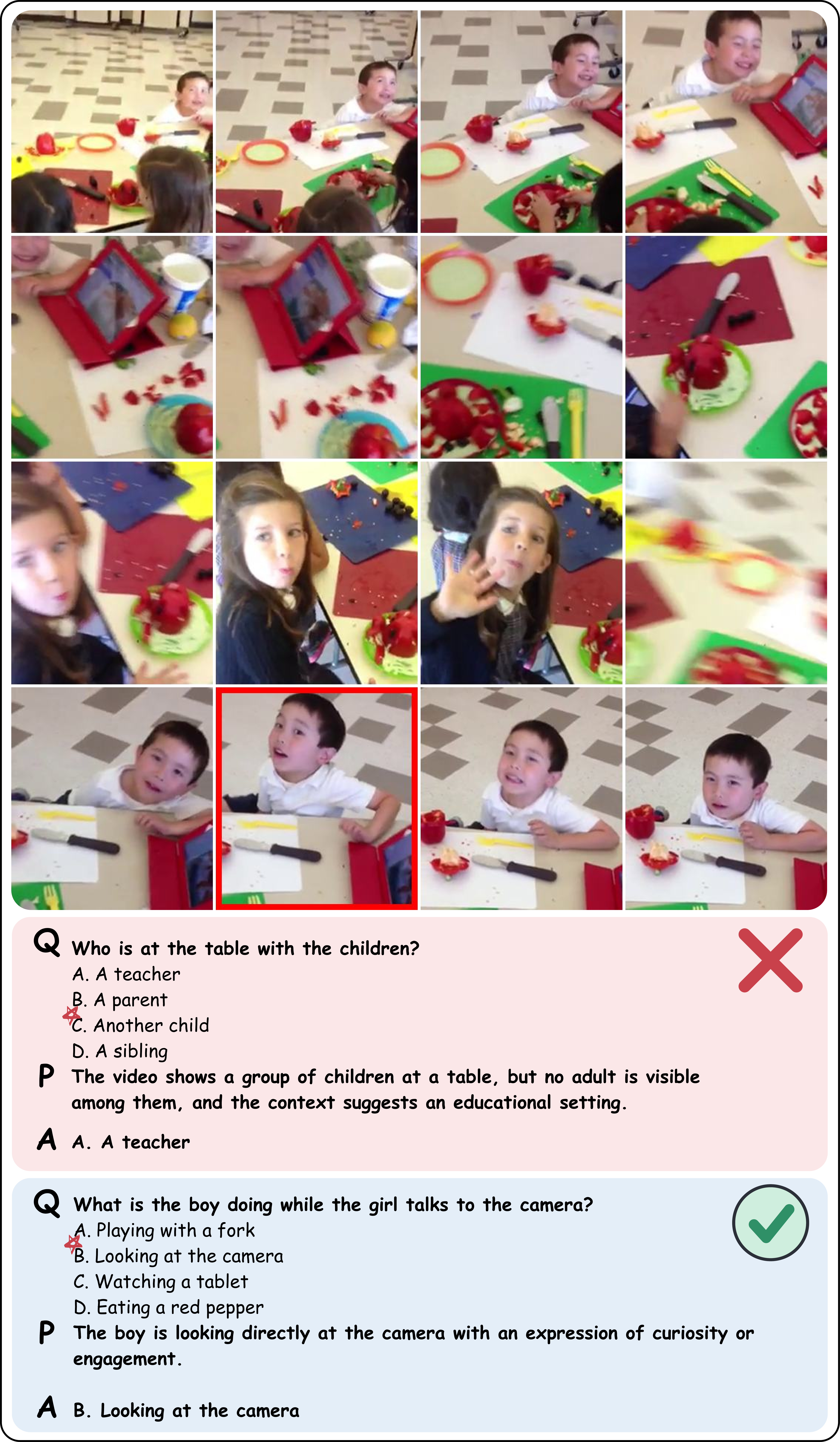}
  \caption{\textbf{Annotations rejected (top) and retained (bottom) by the Teacher-Answer gate on NExT-QA.}}
  \label{fig:qual1_4}
  \vspace{-1em}
\end{figure*}

\Cref{fig:qual1_1,fig:qual1_2,fig:qual1_3,fig:qual1_4} show annotations that the Teacher-Answer gate retains and rejects.
The key frames for each question are marked with red boxes.
For each annotation, we show the question, the pseudo-answer, and the privileged information, and the red-colored star marks the answer confirmed by a human.
An annotation is rejected when an answer of the self-teacher differs from the pseudo-answer, and is retained when all answers reproduce it.
The examples use videos from the training set.

In all four examples, the privileged information of the rejected annotation does not match the video, and the pseudo-answer follows the incorrect privileged information.
The gate rejects an annotation when even one response of the self-teacher differs from the pseudo-answer, so these annotations are rejected.
For example, the self-teacher in \cref{fig:qual1_2} gives a different answer in one of the four responses, although the privileged information states that the person stirs the bowl twice.
The self-teacher receives the video together with the privileged information, and its answer is therefore not determined by the privileged information alone.
The gate can thus reject an annotation with incorrect privileged information, which a check based on the privileged information alone could not detect.
For each video, the annotation that the gate retains has privileged information that matches the video.

\subsection{Annotations of the MLLM Annotator during Training}
\label{sec:app_qual_annot}

\begin{figure*}[th!]
  \centering
  \includegraphics[width=0.8\linewidth]{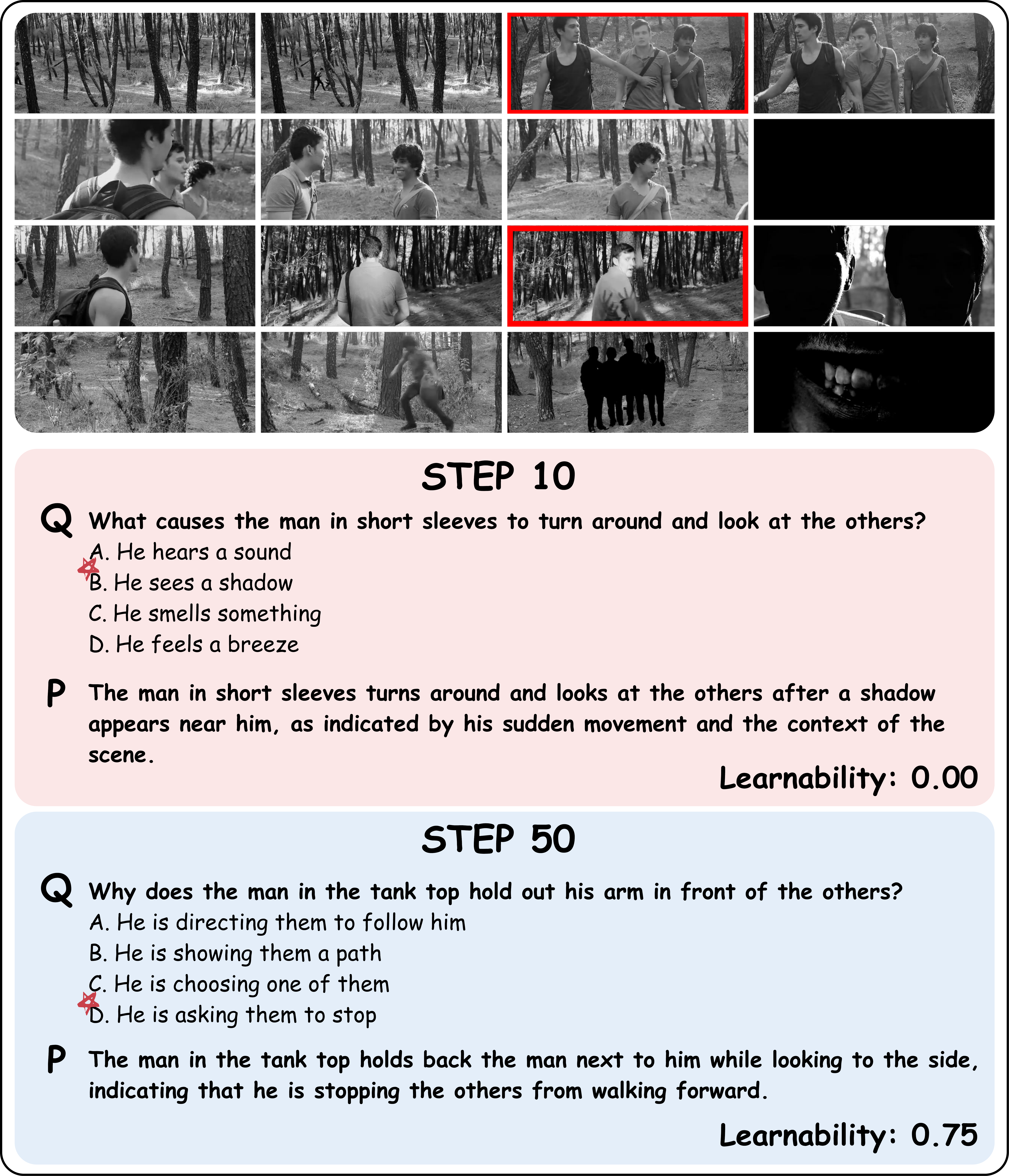}
  \caption{\textbf{Annotations of the MLLM annotator at step 10 (top) and step 50 (bottom) on Video-Holmes.}}
  \label{fig:qual2_1}
  \vspace{-1em}
\end{figure*}

\begin{figure*}[th!]
  \centering
  \includegraphics[width=0.8\linewidth]{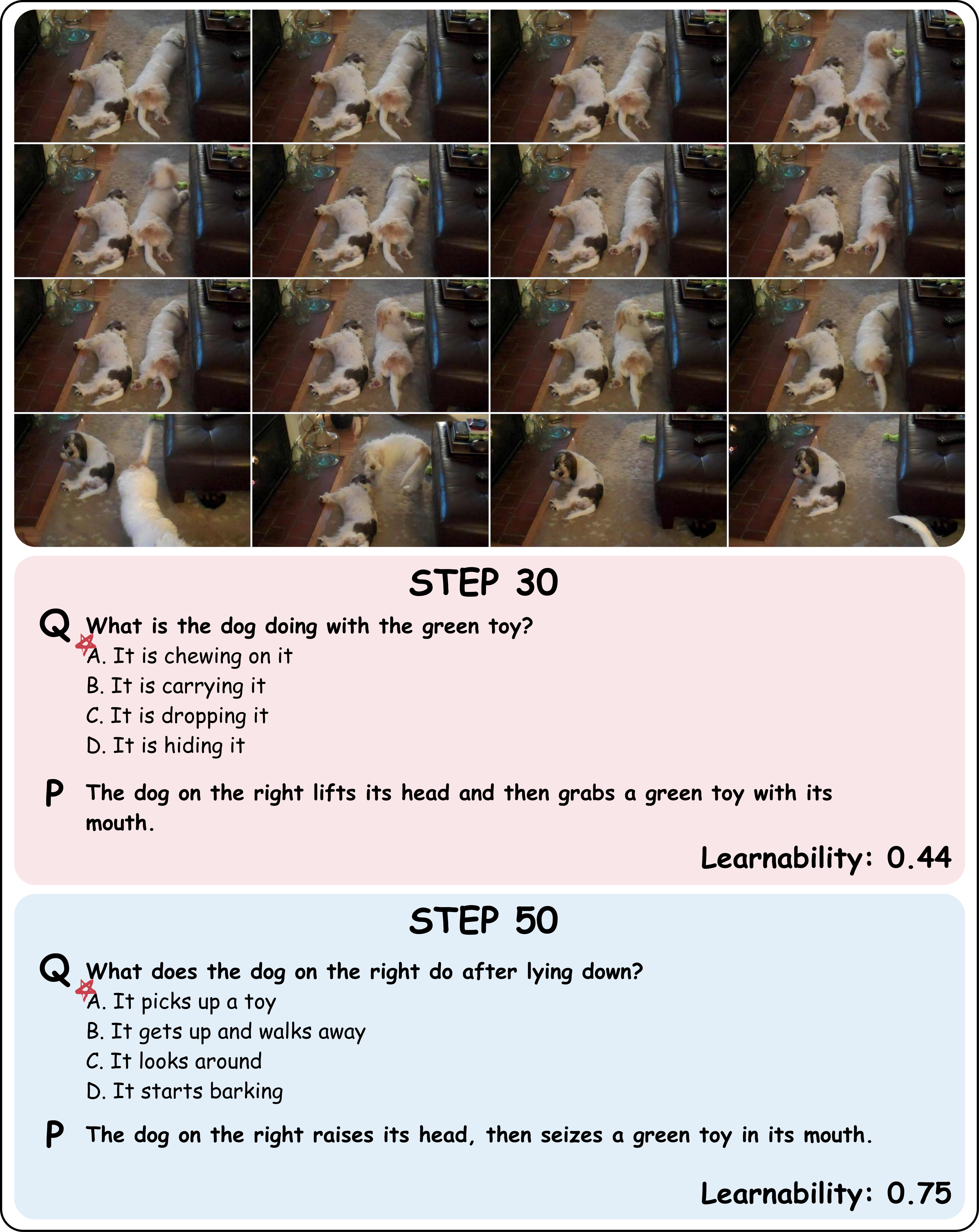}
  \caption{\textbf{Annotations of the MLLM annotator at step 30 (top) and step 50 (bottom) on NExT-QA.}}
  \label{fig:qual2_2}
  \vspace{-1em}
\end{figure*}

\Cref{fig:qual2_1,fig:qual2_2,fig:qual2_3} show the annotations that the MLLM annotator generates for the same video at two training steps.
For each annotation, we show the question, the choices with the pseudo-answer marked by a red star, the privileged information, and the learnability reward (\cref{eq:reward}) measured with the student at the same step.

In \cref{fig:qual2_1}, the annotator at step 10 asks about a cause that is not shown in the video, whereas the annotator at step 50 asks about an action whose purpose is visible in the key frame.
In \cref{fig:qual2_2}, the annotator at step 30 asks what the dog does with the toy, whereas the annotator at step 50 asks about the order of the actions of the dog.
In \cref{fig:qual2_3}, the annotator at step 10 asks about the main activity of the two girls, whereas the annotator at step 50 asks about the position of a man in the background.
% In these examples, the annotator at the later step asks about a more specific part of the video.
In these examples, the annotator at the later step asks about a part of the video that the student answers less consistently.
The learnability reward increases from 0.00 to 0.75, from 0.44 to 0.75, and from 0.00 to 0.94, respectively, where a reward of 0.00 indicates that all rollouts of the student give the same answer.

\subsection{Comparison with the Baselines}
\label{sec:app_qual_cmp}

\begin{figure*}[th!]
  \centering
  \includegraphics[width=0.75\linewidth]{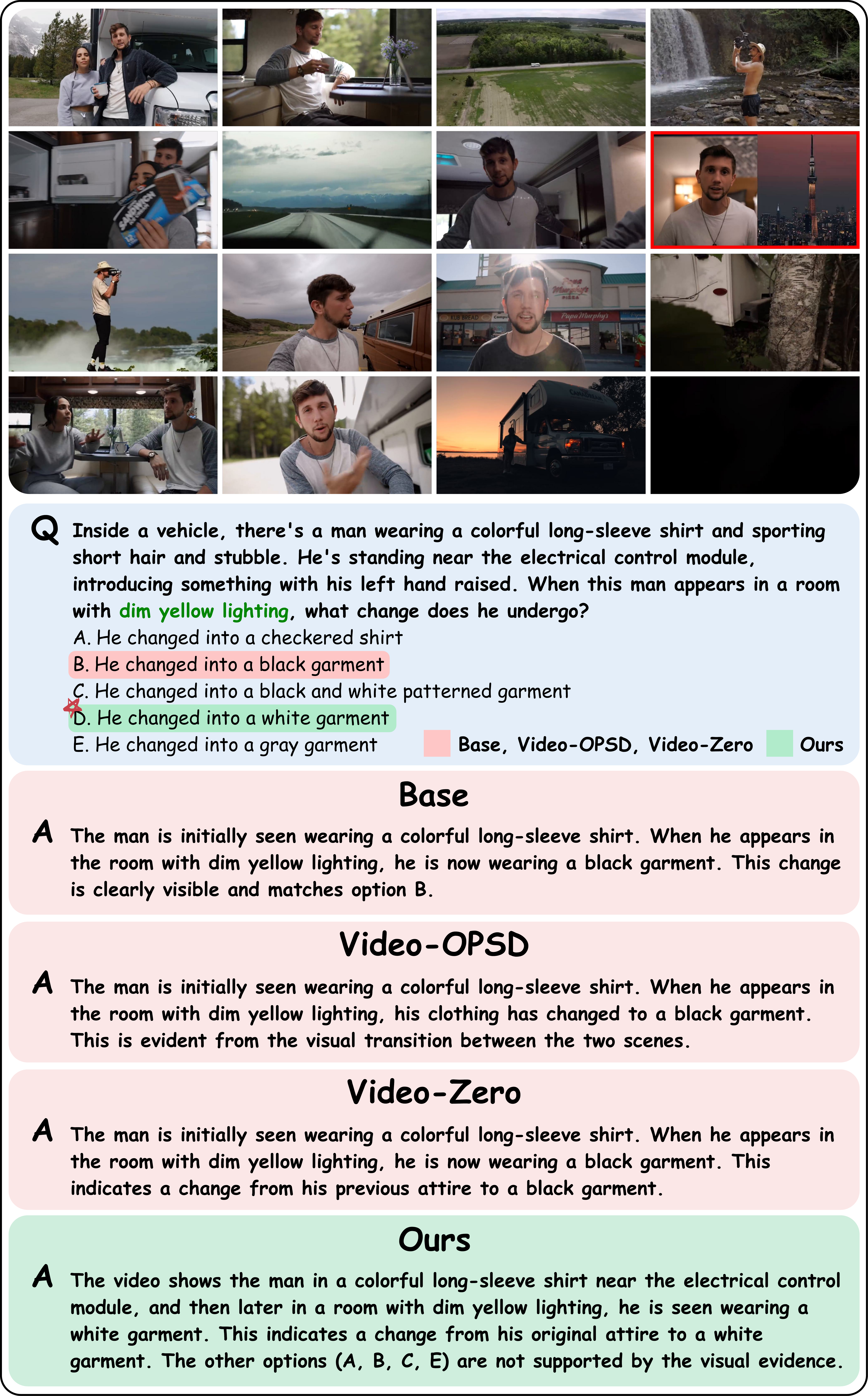}
  \caption{\textbf{Qualitative comparison on LongVideoBench.}}
  \label{fig:qual3_1}
  \vspace{-1em}
\end{figure*}

\begin{figure*}[th!]
  \centering
  \includegraphics[width=0.8\linewidth]{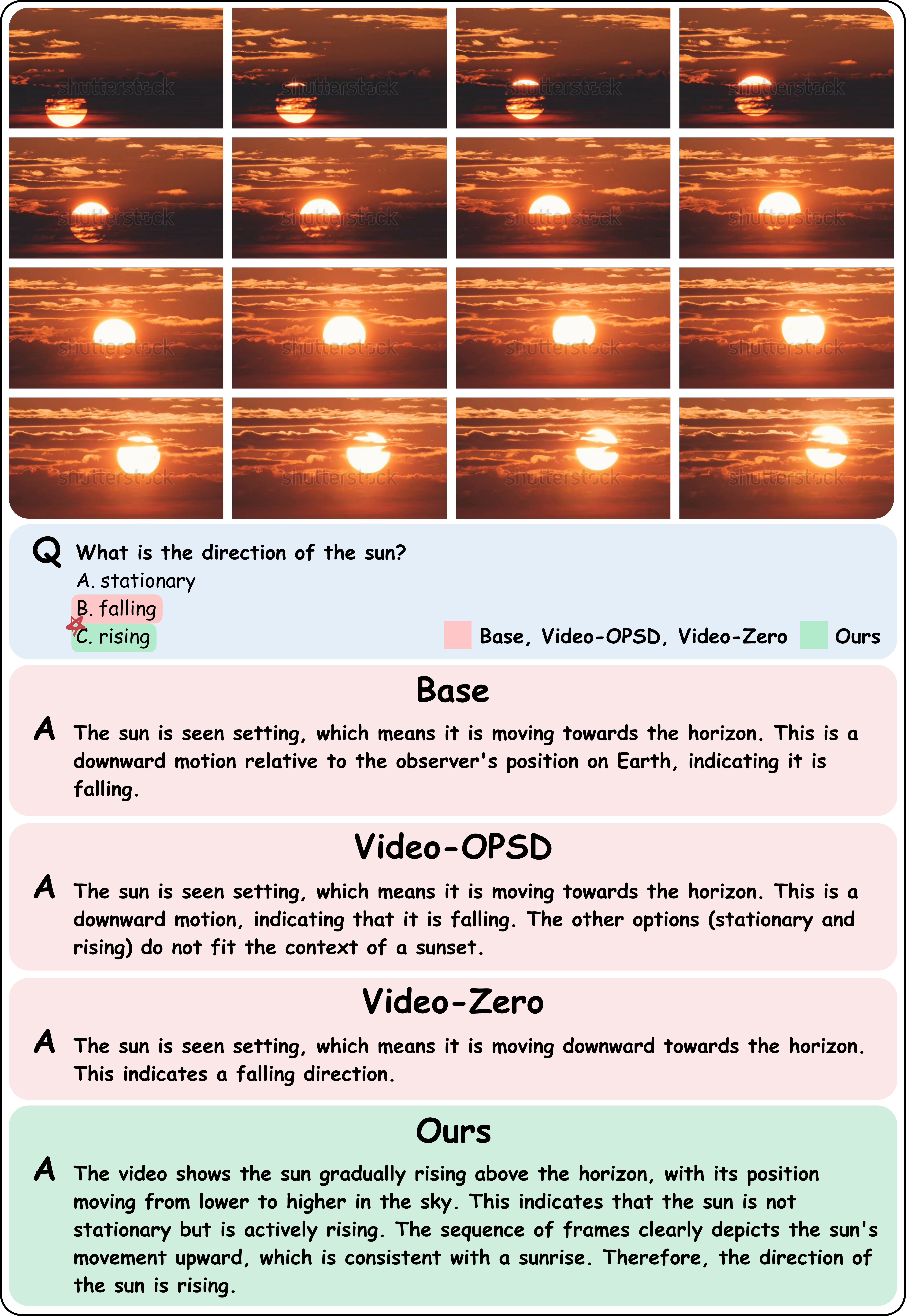}
  \caption{\textbf{Qualitative comparison on TempCompass.}}
  \label{fig:qual3_2}
  \vspace{-1em}
\end{figure*}

\begin{figure*}[th!]
  \centering
  \includegraphics[width=0.8\linewidth]{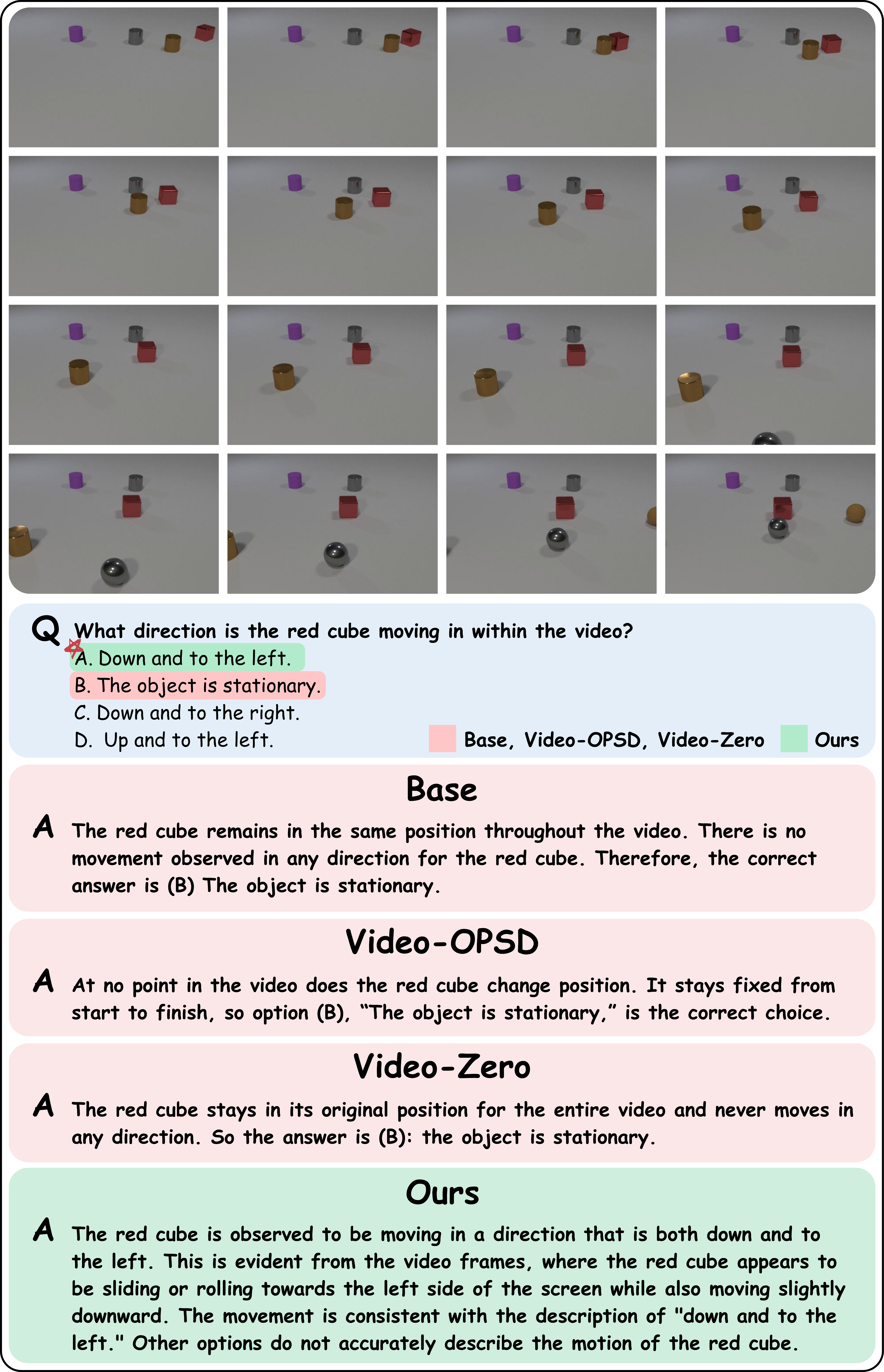}
  \caption{\textbf{Qualitative comparison on MVBench.}}
  \label{fig:qual3_3}
  \vspace{-1em}
\end{figure*}

\Cref{fig:qual3_1,fig:qual3_2,fig:qual3_3} show questions that our method answers correctly and Qwen3-VL-4B, Video-OPSD, and Video-Zero answer incorrectly.
In \cref{fig:qual3_1}, the question asks how the clothing of a man changes in a scene that appears in a single key frame of a long video.
The three baselines answer that the man wears a black garment, whereas our method identifies the white garment in the key frame.
In \cref{fig:qual3_2}, the baselines describe the sun as setting, whereas our method describes the upward movement of the sun across the frames.
In \cref{fig:qual3_3}, the baselines answer that the red cube is stationary, whereas our method describes the movement of the cube toward the lower left.
In these examples, the baselines give answers that are plausible without the video, whereas our method refers to the change that is visible across the frames.

\begin{figure*}[th!]
  \centering
  \includegraphics[width=0.65\linewidth]{Figures/Group_272.pdf}
  \caption{\textbf{Annotations of the MLLM annotator at step 10 (top) and step 50 (bottom) on NExT-QA.}}
  \label{fig:qual2_3}
  \vspace{-1em}
\end{figure*}

\twocolumn[{%
\section{Prompt Setting}
\label{sec:app_prompt}
\begin{promptbox}[before skip=0pt, after skip=12pt]{Prompt of the MLLM annotator}
\ph{video}\\
Watch the video and write one multiple-choice question about it for a video-understanding test, followed by its answer, a short description of what in the video shows the answer, and one sentence on how it shows the answer. The video is \ph{duration} long; each frame is shown with its timestamp.\\
Ask one question of this type:\\
- \ph{type}: \ph{definition}\\
Good questions:\\
- cannot be answered from a single frame or from common sense alone;\\
- have exactly one correct answer that is clearly visible in the video;\\
- use only what is clearly visible: do not invent objects, people, or events;\\
- do not put the answer, or words from it, into the question.\\
P describes, in two or three sentences, who is in the video and what they do in the parts that show the answer, without saying which choice is correct; R then says in one sentence how this shows the answer.\\
The 4 choices must include exactly one correct choice; the others are plausible, similar in length and style, and about things that could happen in this kind of video.\\
Write exactly these five lines and nothing else. Do not write any reasoning, preface, heading, or explanation before or after them:\\
Q: <question>\\
C: <choice 1> | <choice 2> | <choice 3> | <choice 4>\\
A: <the correct choice, copied exactly>\\
P: <two or three sentences on what can be seen in the parts that matter>\\
R: <one sentence on how this shows the answer>\\
Rules for P and R, all of which must hold:\\
- P: describe only what can be seen; do not copy the wording of a choice and do not say which choice is correct;\\
- R: one sentence on how what P describes shows the answer;\\
- keep each on one line.\\
Example of the P and R lines:\\
P: A man in a grey coat sits at a desk in an office. He picks up a phone, looks around and puts it into a drawer.\\
R: So he is hiding the phone.
\end{promptbox}}]

\begin{promptbox}{Prompt of the student}
\ph{video}\\
\ph{question}\\
Options:\\
A. \ph{choice}\\
B. \ph{choice}\\
C. \ph{choice}\\
D. \ph{choice}\\
Answer with the option's letter from the given choices.\\
Provide your thinking process between the <think> and </think> tags, and then give your final answer between the <answer> and </answer> tags.
\end{promptbox}

\begin{promptbox}{Prompt of the self-teacher}
\note{For the Teacher-Answer gate ($N=4$ responses per annotation)}
\ph{video}\\
Notes from another viewer: \ph{P}.\\
\ph{student prompt without the video}
\tcblower
\note{For token-level supervision of the student rollouts}
\ph{video}\\
Notes from another viewer: \ph{P} \ph{R}.\\
\ph{student prompt without the video}
\end{promptbox}

\newpage